\documentclass{IEEEtran}

\usepackage[pdftex]{graphicx}  % {.pdf,.jpeg,.png}
\graphicspath{{./figures/}}
\DeclareGraphicsExtensions{.pdf,.jpeg,.png,.jpg}
\usepackage{float}
\usepackage{subfigure}
\usepackage{textcomp}
\usepackage{stfloats}
\usepackage{url}
\usepackage{verbatim}
\usepackage{amsmath}  % MATH PACKAGES
\usepackage{amsfonts}
\usepackage{amssymb}  % MATH PACKAGES
\usepackage{mathrsfs} % MATH PACKAGES
\usepackage{dutchcal}

\usepackage{array}    % ALIGNMENT PACKAGES
\usepackage{booktabs}
\usepackage{multirow}
\usepackage{multicol}
\usepackage{tabularx}
\usepackage[ruled,linesnumbered]{algorithm2e}

\usepackage{amsthm}   % MATH PACKAGES
\theoremstyle{definition}  % environment of definition
\usepackage[nocompress]{cite}  % CITATION PACKAGES

\usepackage{color}
\usepackage{bm}  % bold font in math
\usepackage{enumitem}  % style of enumerate item
\begin{document}
\title{Fuzzy Cluster-Aware Contrastive Clustering for Time Series}
\author{Congyu Wang, Mingjing Du, Xiang Jiang and Yongquan Dong% <-this % stops a space
    \IEEEcompsocitemizethanks{
        \IEEEcompsocthanksitem This work is supported by the National Natural Science Foundation of China (No. 62006104), Postgraduate Research \& Practice Innovation Program of Jiangsu Normal University (No. 2024XKT2583).\textit{(Corresponding author: Mingjing Du and Yongquan Dong.)}
        \IEEEcompsocthanksitem C. Wang, M. Du, X. Jiang and Yongquan Dong are with the Jiangsu Key Laboratory of Educational Intelligent Technology, School of Computer Science and Technology, Jiangsu Normal University, Xuzhou, 221116, China. (e-mail: wangcongyu@jsnu.edu.cn; dumj@jsnu.edu.cn; xjiang@jsnu.edu.cn; tomdyq@163.com).
    }% 
}

% The paper headers
\markboth{IEEE TRANSACTIONS ON NEURAL NETWORKS AND LEARNING SYSTEMS, ~Vol.~xx, No.~xx, 2023}
{Sun \MakeLowercase{\textit{et al.}}: Bare Demo of IEEEtran.cls for Computer Society Journals}

\IEEEtitleabstractindextext{
    \begin{abstract}
    % The rapid growth of unlabeled time series data, driven by the Internet of Things (IoT), poses significant challenges in uncovering underlying patterns. Traditional unsupervised clustering methods often fail to capture the complex nature of time series data. Recent deep learning-based clustering approaches, while effective, struggle with insufficient representation learning and the integration of clustering objectives. To address these issues, we propose a fuzzy cluster-aware contrastive clustering framework (FCACC) that jointly optimizes representation learning and clustering. Our approach introduces a novel three-view data augmentation strategy to enhance feature extraction and a cluster-aware hard negative sample generation mechanism to improve discriminative power. By leveraging fuzzy clustering, FCACC dynamically generates cluster structures to guide the contrastive learning process, resulting in more accurate clustering. Extensive experiments on 40 benchmark datasets show that FCACC outperforms the selected baseline methods (eight in total), providing an effective solution for unsupervised time series learning.
    
    The rapid growth of unlabeled time series data, driven by the Internet of Things (IoT), poses significant challenges in uncovering underlying patterns. Traditional unsupervised clustering methods often fail to capture the complex nature of time series data. Recent deep learning-based clustering approaches, while effective, struggle with insufficient representation learning and the integration of clustering objectives. To address these issues, we propose a fuzzy cluster-aware contrastive clustering framework (FCACC) that jointly optimizes representation learning and clustering. 
    Our approach introduces a novel three-view data augmentation strategy to enhance feature extraction by leveraging various characteristics of time series data. Additionally, we propose a cluster-aware hard negative sample generation mechanism that dynamically constructs high-quality negative samples using clustering structure information, thereby improving the model’s discriminative ability. 
    By leveraging fuzzy clustering, FCACC dynamically generates cluster structures to guide the contrastive learning process, resulting in more accurate clustering. Extensive experiments on 40 benchmark datasets show that FCACC outperforms the selected baseline methods (eight in total), providing an effective solution for unsupervised time series learning.

    \end{abstract}

}

% make the title area
\maketitle
\IEEEdisplaynontitleabstractindextext
\IEEEpeerreviewmaketitle

% import sections one by one
\section{Introduction}
\label{sec:introduction}
\IEEEPARstart The widespread adoption of sensors and advancements in Internet of Things (IoT) technologies have led to a surge in time series data generated by automated devices in daily life\cite{yang2023unsupervised}. However, labeling this data requires specialized expertise and involves substantial resource investment, resulting in a scarcity of labeled data\cite{qian2021weakly}. This shortage hinders the effectiveness of traditional supervised learning methods\cite{meng2023mhccl}. In contrast, clustering, as a powerful unsupervised learning technique\cite{saxena2017review}\cite{xue2023fast}\cite{zhao2018icfs}, has emerged as a widely used alternative by uncovering hidden patterns within unlabeled time series data\cite{he2022soft}\cite{li2021contrastive}.

As time series analysis evolves, there has been a growing interest in applying deep learning techniques to time series clustering tasks\cite{zhang2024self}. In particular, contrastive learning, a deep self-supervised learning paradigm, has emerged as an effective technique for capturing intrinsic patterns in data\cite{li2022twin}\cite{liu2024timesurl }\cite{yue2022ts2vec}. Contrastive learning-based time series clustering methods automatically extract latent features from raw time series data by constructing positive and negative sample pairs\cite{zhong2023deep}. These methods demonstrate superior performance in handling complex, noisy, and long-sequence data compared to traditional techniques\cite{peng2024cross}. Despite these advancements, existing methods still suffer from limited representation ability and insufficient joint optimization.

Firstly, existing contrastive learning methods predominantly employ single augmentation strategies (e.g., sub-series cropping\cite{franceschi2019unsupervised} for short-term dependencies or transformation-based approaches\cite{eldele2021time} for long-term patterns), resulting in incomplete feature capture.
Moreover, traditional methods treat augmentations of other instances as negative samples\cite{yue2022ts2vec} and maximize the distance between anchor points and negative samples. This forces the anchor to diverge significantly from other instances, potentially conflicting with clustering objectives.

Furthermore, in the integration of representation learning and clustering, current deep clustering methods treat representation learning and clustering as separate components, increasing complexity in parameter tuning and causing inconsistencies in optimization objectives\cite{li2024feature}\cite{ma2020self}.
Although some methods combine these objectives by adding their losses\cite{zhong2023deep}, they inadequately integrate the intrinsic relationship between representation learning and clustering. 
This additive approach overlooks both the guiding role of clustering information in feature extraction and the feedback from feature representation during clustering, ultimately constraining overall performance.

To address the above issues, this paper proposes a fuzzy cluster-aware contrastive clustering framework for time series, called FCACC.
In representation learning, we propose a three-view data augmentation strategy based on multiple cropping and perturbation to comprehensively capture time series characteristics. 
Moreover, we introduce two strategies built on the cluster-awareness generation module for hard negative sample generation and positive-negative sample pair selection, thereby producing cluster-friendly representations. In clustering, we employ the fuzzy c-means (FCM) algorithm, which better handles the complex structure of time series data. 
Finally, a joint optimization mechanism is designed for representation learning and clustering. 
We incorporate the fuzzy membership values from FCM into the cluster-awareness generation module, using cluster structure information to guide positive-negative sample selection in contrastive learning. This ensures that samples within the same cluster share more similar representations, further optimizing clustering results. This mechanism enables iterative improvement between the FCM-based clustering and contrastive learning representations. In summary, the main contributions of this work are as follows:
\begin{itemize}
    \item We introduce a three-view data augmentation strategy based on multiple cropping and perturbation, effectively leveraging various characteristics of time series to enhance the model's feature extraction ability.
    \item We propose cluster-aware contrastive learning, where fuzzy membership derived from FCM dynamically regulates sample selection, facilitating simultaneous feature refinement and cluster structure optimization. This approach enables the learning of cluster-friendly representations.
    \item We design a hard negative sample generation strategy based on cluster awareness to construct high-quality negative samples to boost the model’s discriminative power.
\end{itemize}
\section{Related Work}
\label{sec:related_work}
In this section, we review contrastive learning algorithms for time series, as well as deep time series clustering methods closely related to this study.
\subsection{Contrastive Learning}
In recent years, the application of contrastive learning in time series analysis has received widespread attention\cite{peng2024cross}\cite{wang2023contrast}\cite{poppelbaum2022contrastive}. 
Learning invariant feature representations through data augmentation is a common approach in time series contrastive learning, which has demonstrated outstanding performance in unsupervised tasks\cite{chang2024timedrl}\cite{luo2023time}. 
Data augmentation is a critical component of this approach\cite{luo2023time}\cite{kim2023contrastive}, and various strategies have been proposed to enhance the model’s representational capabilities. 

Existing time series augmentation strategies can be mainly categorized into two types: transformation-based augmentation and cropping-based augmentation. Transformation-based augmentation\cite{eldele2021time} leverages the transformation consistency of time series to generate transformed views through operations such as scaling, rearranging, and perturbing, encouraging the model to capture features that are invariant to transformations. On the other hand, cropping-based augmentation generates views by cropping different segments of the time series. Tonekaboni et al.\cite{tonekaboni2021unsupervised} enhance the local smoothness of representations by learning temporal consistency from adjacent segments generated through cropping; Franceschi et al.\cite{franceschi2019unsupervised} use subseries consistency to make the representation of the time series closer to its sampled subseries. Furthermore, TS2Vec\cite{yue2022ts2vec} improves the robustness of time series representations by combining contextual consistency and forcing each timestamp to reconstruct itself in different contexts.

In addition to continuous improvements in augmentation strategies, the research path of time series contrastive learning has also expanded in other areas. For example, TimesURL\cite{liu2024timesurl} introduces a hard negative sample generation strategy by mixing positive and negative samples in both instance-level and time-level losses. It uses batch data point indices as pseudo-labels to generate more challenging sample pairs, effectively increasing the difficulty of contrastive learning and enhancing the model's feature discriminative ability. On the other hand, CDCC\cite{peng2024cross} combines time-domain and frequency-domain information to optimize the alignment of cross-domain latent representations, while also improving clustering performance.

Compared to existing methods, our approach exhibits significant differences in the key technical design of contrastive learning. First, unlike the single augmentation strategy and traditional dual-view architecture, we integrate contextual consistency, subseries consistency, and transformation consistency through a design based on multiple cropping and perturbation, generating diverse and interrelated augmented views. Second, in contrast to existing methods, we combine cluster structures to generate higher-quality hard negative samples at both the instance level and time level, effectively reducing the interference of false negative samples and thereby enhancing the model's feature discriminative ability.

\subsection{Deep Time Series Clustering}
Deep time series clustering is one of the key methods in unsupervised data mining. Existing methods can be broadly categorized into separated optimization methods and joint optimization methods\cite{alqahtani2021deep}.

The separated optimization methods typically follow a "representation first, then clustering" strategy. These methods use deep learning models to extract feature representations from time series data, and then apply traditional clustering algorithms (such as K-means) to perform clustering tasks. For example, Chen et al.\cite{chen2022clustering} use an RNN to encode time series data and perform clustering in a unified latent space; T-LSTM\cite{baytas2017patient}, designed specifically for medical time series, uses an improved LSTM model to capture both short-term and long-term memory dependencies, providing efficient feature representations for patient subtyping. However, the main issue with these methods is that feature extraction and clustering are treated separately. Clustering cannot provide feedback to guide the feature extraction process, and the complexity of parameter tuning further limits the model's performance\cite{ma2020self}.

To overcome these issues, researchers have gradually shifted towards joint optimization methods, which integrate feature extraction and clustering objectives into a single optimization framework. DTC\cite{madiraju2018deep} combines autoencoder-based dimensionality reduction with clustering objectives, exploring joint clustering for time series data. STCN\cite{ma2020self} introduces a self-supervised learning framework that combines dynamic feature extraction with clustering, iteratively optimizing pseudo-labels and model parameters to enhance clustering accuracy. TCGAN\cite{huang2023tcgan} introduces generative adversarial networks (GANs) to learn latent distributions through adversarial training, enhancing performance in complex unlabeled scenarios. Meanwhile, Zhong et al.\cite{zhong2023deep} propose a deep time contrastive clustering, which combines contrastive learning with K-means to jointly optimize feature representations and clustering performance.

Unlike existing methods that merely combine feature representations and clustering objectives, our method introduces a cluster-awareness generation module, which serves as a collaborative mechanism between contrastive learning and clustering. This module allows the clustering results to guide the generation of cluster-friendly representations, while the enhanced representations, in turn, refine the clustering process, creating a mutually reinforcing cycle between representation learning and clustering.
\section{Proposed Algorithm}
\label{sec:algorithm}
\begin{figure*}[ht]
    \centering
    \includegraphics[width=\linewidth]{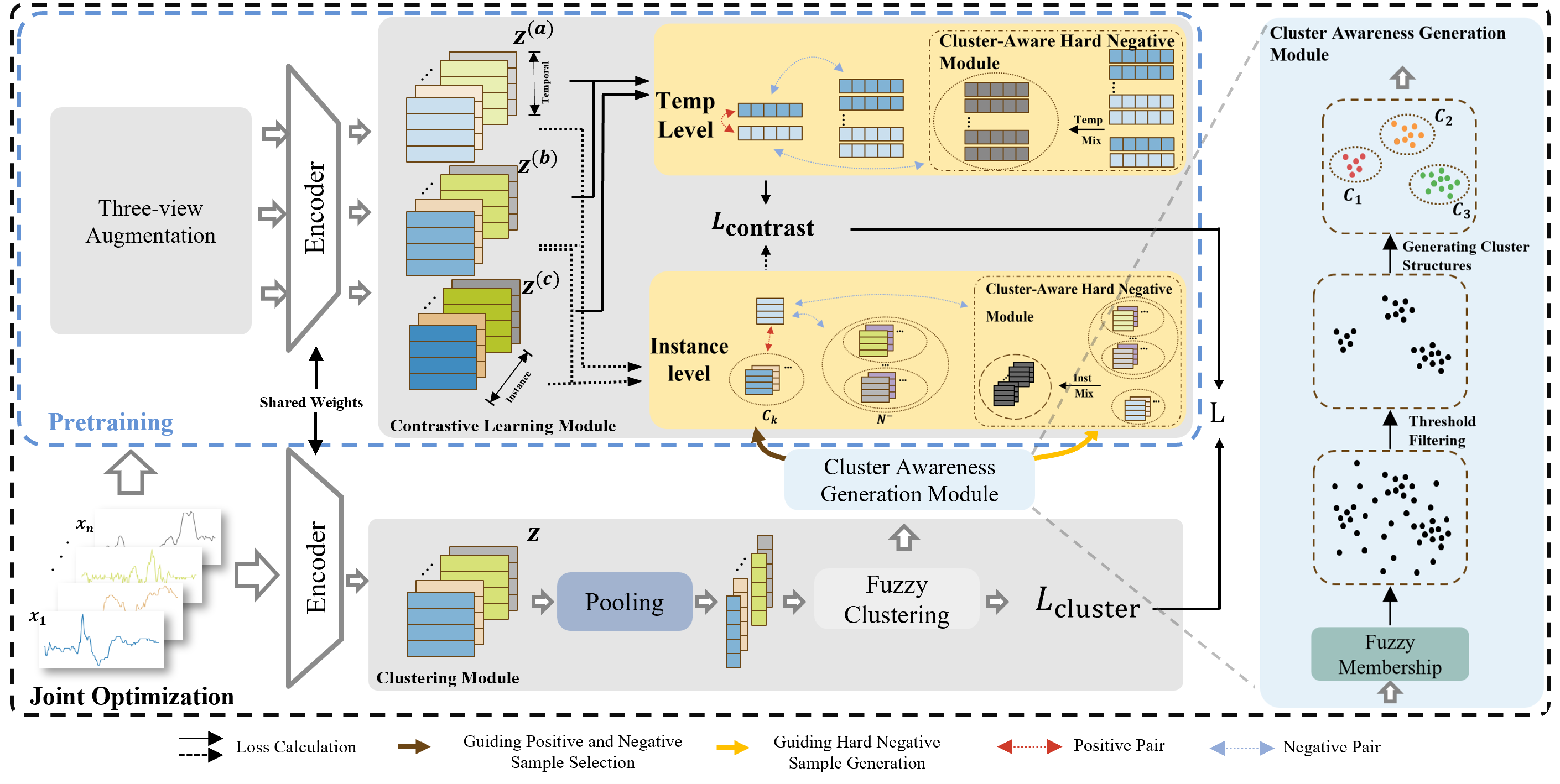}
    \caption{Overall Structure of FCACC Framework. This diagram illustrates the complete FCACC framework, which is organized into a pre-training stage and a joint optimization stage. The system comprises three primary modules: (1) the contrastive learning module, which enhances representation quality via multi-view augmentation and hard negative sample generation; (2) the fuzzy clustering module, which effectively models complex membership relationships in the data; and (3) the cluster-awareness generation module, which dynamically constructs cluster structures that steer the joint optimization process. The directional arrows denote the data flow among the modules, emphasizing their collaborative interaction through the cluster-awareness mechanism.}
    \label{fig:framework}
\end{figure*}
% \subsection{Overview}
% \label{subsec:overview}
This paper proposes a fuzzy cluster-aware time series contrastive clustering framework (FCACC), as shown in Figure 1. 
The framework integrates three core modules: the contrastive learn module, the fuzzy clustering module, and the cluster-awareness generation module and operates in two stages: pretraining and joint optimization.
In the pre-training stage, the contrastive learning module is employed along with multi-view data augmentation and hard negative sample generation to learning robust time series representations for subsequent optimization. 
In the joint optimization stage, the fuzzy clustering module generates fuzzy membership degrees to capture complex data patterns, while the cluster-awareness generation module dynamically constructs the core cluster structure and injects it into the contrastive learning module. 
This dynamic guidance drives both the selection of positive and negative samples and the generation of hard negative samples, ultimately enabling a deep, joint optimization of representation learning and clustering objectives. 
The design and function of each module will be elaborated in detail in the following sections.
\subsection{Contrastive Learning Module}
\label{subsec:contrastive_learning_module}
\subsubsection{Three-View Data Augmentation Strategy Based on Multiple Cropping and Perturbation}
\label{subsubsec:data_arg}
\begin{figure}[ht]
    \centering
    \includegraphics[width=1\linewidth]{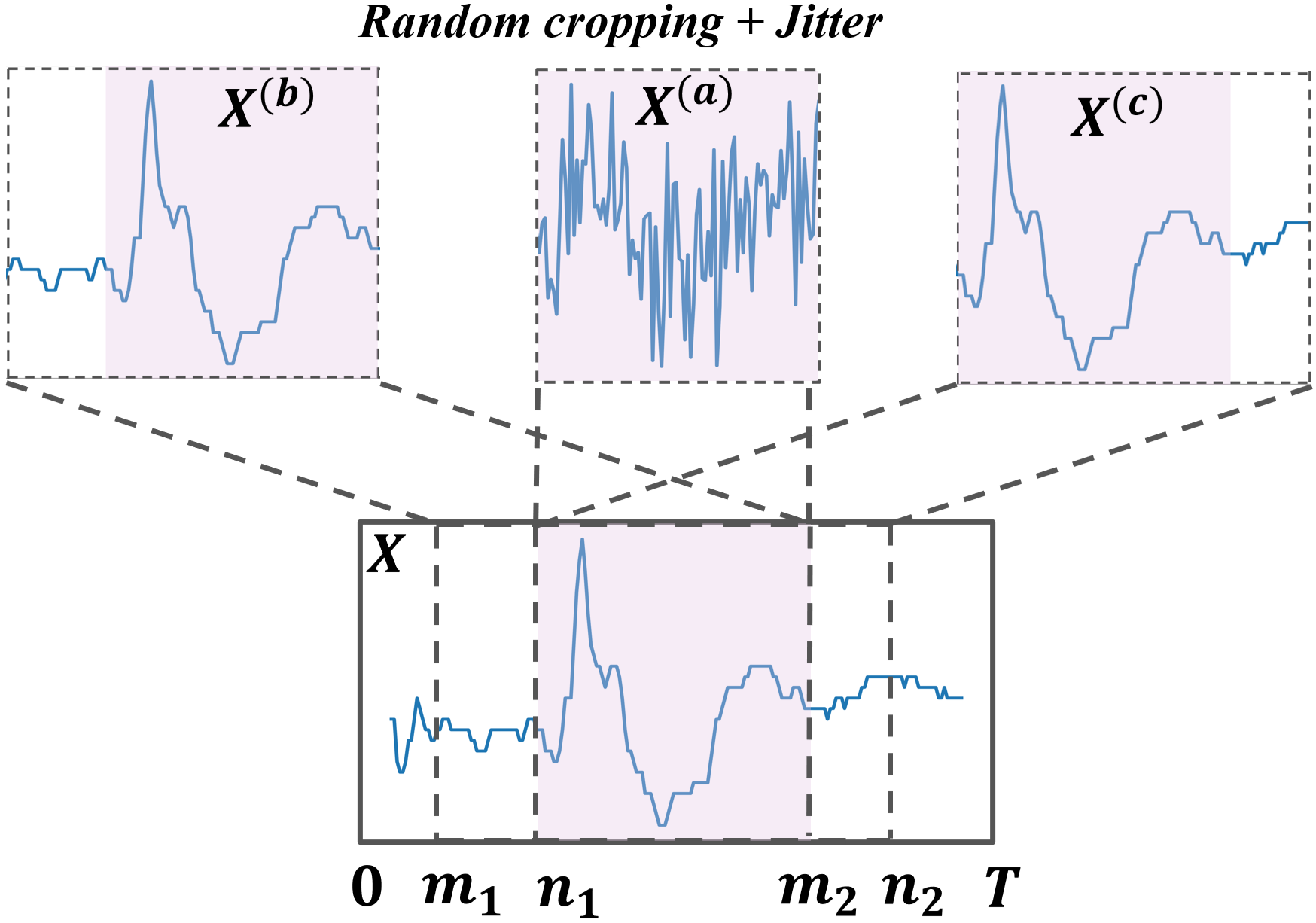}
    \caption{Data Augmentation Process. The input raw time series is processed through multiple random cropping operations to generate three subsequences as shown in the figure. These subsequences share the same overlapping region \([n_1, m_2]\). Among them, the first time segment \([n_1, m_2]\) undergoes additional perturbations to form \( \boldsymbol{X}^{(a)} \), while the latter two segments remain unchanged, forming \( \boldsymbol{X}^{(b)} \) and \( \boldsymbol{X}^{(c)} \), respectively.
}
    \label{fig:enter-label}
\end{figure}
We propose a three-view data augmentation strategy to enhance the feature representation of time series data in contrastive learning. Unlike traditional approaches that rely on a single strategy or two-view architecture—such as transformation-based or cropping techniques, which capture only specific consistencies (e.g., context, subseries, or transformation consistency)—our strategy integrates multiple cropping and perturbation techniques to comprehensively capture various characteristics of time series data.  

As shown in Figure 2, this strategy constructs three augmented views, \( \boldsymbol{X}^{(a)} \), \( \boldsymbol{X}^{(b)} \), and \( \boldsymbol{X}^{(c)} \), from the time series \( \boldsymbol{X} \) to comprehensively capture various characteristics of time series data. Specifically, we randomly select three time segments, \([m_1, m_2]\), \([n_1, m_2]\), and \([n_1, n_2]\), which have distinct contexts but share the overlapping region \([n_1, m_2]\), ensuring that \( 0 < m_1 < n_1 \leq m_2 \leq n_2 \leq T \). Among them, the first segment \([n_1, m_2]\) undergoes perturbation to form \( \boldsymbol{X}^{(a)} \), while the latter two remain unchanged as \( \boldsymbol{X}^{(b)} \) and \( \boldsymbol{X}^{(c)} \), respectively.  

The augmented data is processed by a shared-weight encoder. In the proposed framework, we employ a backbone network with triple shared weights. 
Specifically, the three augmented views share the same backbone encoder \( f \), which extracts their representations, denoted as:  \( \boldsymbol{Z}^{(a)} = f(\boldsymbol{X}^{(a)}),  \boldsymbol{Z}^{(b)} = f(\boldsymbol{X}^{(b)}), \boldsymbol{Z}^{(c)} = f(\boldsymbol{X}^{(c)}) \).  

In contrastive learning, pairing augmented views can enforce consistency at different levels. By pairing \( (\boldsymbol{Z}^{(a)}, \boldsymbol{Z}^{(b)}) \), the framework enhances both transformation and subseries consistency, thereby improving robustness to noise and dynamic variations as well as local feature extraction. 
Meanwhile, pairing \( (\boldsymbol{Z}^{(b)}, \boldsymbol{Z}^{(c)}) \), which are temporally adjacent and share the overlapping region \([n_1, m_2]\), promotes contextual consistency and better captures global dependencies. 
Thus, this three-view augmentation strategy effectively leverages diverse sources of variation while preserving view correlations, significantly enhancing feature learning and representation quality.

\subsubsection{Cluster-Aware Hard Negative Sample Generation Strategy}
% \label{subsubsec:data_arg}
% \begin{figure*}[ht]
%     \centering
%     \includegraphics[width=1\linewidth]{figures/hard_native.png}
%     \caption{Comparison of Cluster-Aware Hard Negative Sample Generation and Its Effect. The left figure demonstrates the generation of Clustering Universum negative samples (in blue) by proportionally mixing time series from different clusters using Mixup. These samples do not belong to any original cluster but are close to the boundary of the positive sample region, thereby providing more challenging learning signals. The right figure provides an example using the partial t-SNE visualization of the GesturePebbleZ1 training set from the UCR archive, showing the distribution of samples in the embedding space: For each positive anchor (red square), the original negative samples (gray squares) contain many easy negative samples and a few same-cluster samples. Mixing same-cluster positive samples generates incorrect hard negative samples (green triangles). In contrast, Clustering Universum negative samples (blue triangles) avoid the misclassification of same-class samples as negative samples.}
%     \label{fig:enter-label}
% \end{figure*}
\begin{figure*}[!ht]
    \centering
    \subfigure[]{
        \centering
        \includegraphics[width=0.59\linewidth]{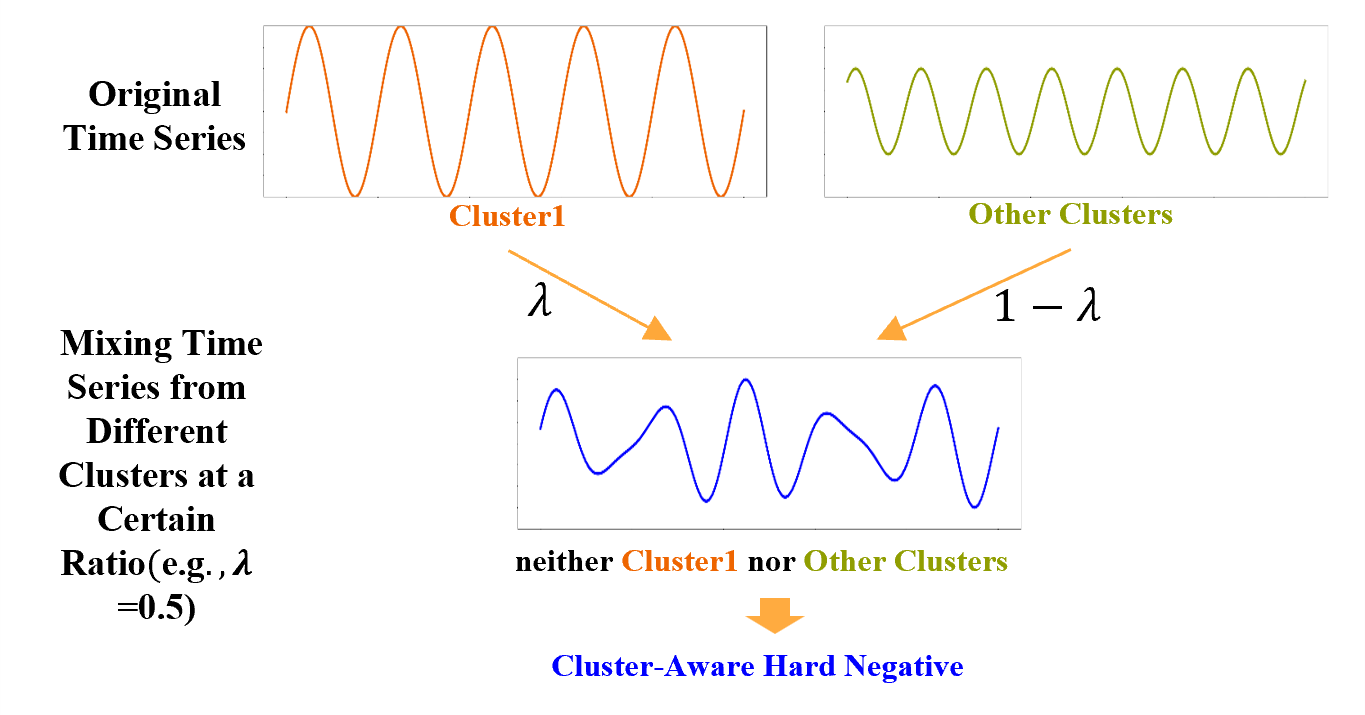}  % 设置子图宽度为33%
        \label{subfig:hard_a}
    }
    \subfigure[]{
        \centering
        \includegraphics[width=0.37\linewidth]{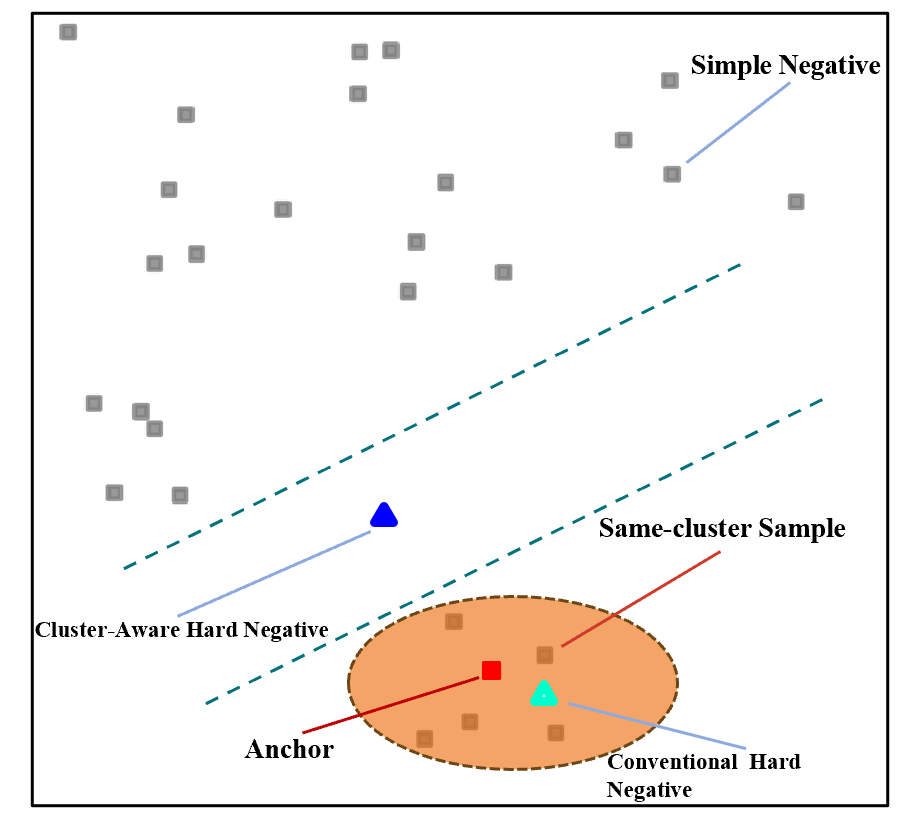}  % 设置子图宽度为33%
        \label{subfig:hard_b}
    }
    \caption{Cluster-aware Hard Negative Sample Generation and its Effect Comparison. The \ref{sub@subfig:hard_a} illustrates the process of generating cluster-aware hard negative samples. We generate these samples by mixing time series from different clusters at a certain ratio. These samples do not belong to any original cluster but are close to the boundary of the positive sample region, thus providing more challenging learning signals. The \ref{sub@subfig:hard_b} shows an example of the distribution of samples in the embedding space from a portion of the GesturePebbleZ1 training set in the UCR archive. For each positive anchor (red square), the original negative samples (gray squares) include many easy negative samples (gray squares far from the positive samples) and a few same-cluster samples (gray squares close to the positive sample). Mixing positive samples from the same cluster results in hard negative samples (green triangles) that are very similar to the positive samples. In contrast, cluster-aware hard negative samples (blue triangles) avoid misclassifying same-cluster positive samples as negative samples.
    }
    % \label{fig:example} %\ref{subfig:parameter_c}
\end{figure*}

After extracting the representations through the encoder, we obtain the high-dimensional latent representations of the time series. Next, the contrastive learning approach selects positive and negative sample pairs from these representations to optimize the model. However, due to the unique local smoothness and Markovian properties of time series data, most of the selected negative samples are ``simple negative sample'', and the difference between these negative samples and the positive samples is too large, providing small gradient contributions to the model's training, thus limiting the performance of contrastive learning\cite{liu2024timesurl}.

To address the limitations of simple negative samples, some studies have attempted to introduce hard negative sample generation strategies to enhance the model's discriminative power by reducing the difference between positive and negative samples. 
Traditional Universum-style mechanisms generate generalized negative samples that “do not belong to any original class” (e.g., an animal that is half dog and half cat, thus neither a dog nor a cat), thereby increasing the diversity and difficulty of negative samples. However, in unsupervised scenarios, these mechanisms often rely on batch sample indices as pseudo-labels\cite{kalantidis2020hard}, ignoring the semantic relationships between samples. This design risks misclassifying same-cluster samples as negative samples, introducing false negatives that degrade the performance of contrastive learning.

The impact of such mislabeling on model performance is visually demonstrated in Figure 3(b). In the GesturePebbleZ1 dataset from the UCR archive, each  positive sample has corresponding negative samples, including both simple negative samples and same-cluster samples. Simple negative samples, being far from the positive sample, contribute little to the contrastive loss. In contrast, same-cluster samples, when mislabeled as negative, can generate incorrect hard negatives. These false negatives, which are closer to the positive sample than the actual same-cluster samples, undermine the effectiveness of the hard negative sample generation strategy and negatively affect the model’s learning process.

Therefore, we propose a cluster-aware hard negative sample generation strategy. The cluster structure derived from clustering is applied in the mixed generation process of samples to ensure that generated negative samples are semantically distinguishable, thereby avoiding the misuse of same-cluster samples as negatives. 

Building on this idea, we design a dual Universum approach to generate high-quality hard negative samples at both the time and instance levels. At the time level, we mix features from different timestamps to generate new negative samples, providing richer learning signals to the model. At the instance level, we mix positive samples from the anchor’s cluster with selected negative samples to create “universal negative samples” that do not belong to any cluster, as shown in Figure 3(a). This dual view-based design generates high-quality hard negatives in the embedding space, improving the model's ability to distinguish complex samples and enhancing the performance of contrastive learning.

To further illustrate the implementation of the dual Universum approach, we provide its specific mathematical representation.  Let \( i \) denote the index of the input time series, and \( t \) denote the timestamp. The representations \( \mathbf{z}^{(a)}_{i,t} \), \( \mathbf{z}^{(b)}_{i,t} \), and \( \mathbf{z}^{(c)}_{i,t} \) correspond to the three augmented views of \( \mathbf{x}_i \) at timestamp \( t \). 

The time-level mixed universe for the \( i \)-th time series at timestamp \( t \)  can be expressed as:

\begin{equation}
    % \label{eq_1}
    \begin{aligned}
        \hat{h}_{i,t}^{(a)} &= \lambda_{1}\cdot z_{i,t}^{(a)} + (1-\lambda_{1})\cdot z_{i,t\prime}^{(a)} \\
        \hat{h}_{i,t}^{(b)} &= \lambda_{1}\cdot z_{i,t}^{(b)} + (1-\lambda_{1})\cdot z_{i,t\prime}^{(b)} \\
        \hat{h}_{i,t}^{(c)} &= \lambda_{1}\cdot z_{i,t}^{(c)} + (1-\lambda_{1})\cdot z_{i,t\prime}^{(c)} 
    \end{aligned}
\end{equation}
where \( t' \) is randomly selected from the set of timestamps \( \Omega \) within the overlapping range of the two subsequences, with \( t' \neq t \). 

Similarly, the instance-level mixed universe indexed by \( (i, t) \) can be expressed as:
\begin{equation}
    % \label{eq_1}
    \begin{aligned}
         & \tilde{h}_{i,t}^{(a)}=\lambda_{2}\cdot z_{k,t}^{(a)}+(1-\lambda_{2})\cdot z_{j,t}^{(a)} \\
         & \tilde{h}_{i,t}^{(b)}=\lambda_{2}\cdot z_{k,t}^{(b)}+(1-\lambda_{2})\cdot z_{j,t}^{(b)} \\
         & \tilde{h}_{i,t}^{(c)}=\lambda_{2}\cdot z_{k,t}^{(c)}+(1-\lambda_{2})\cdot z_{j,t}^{(c)}
    \end{aligned}
\end{equation}
where \( \lambda_1, \lambda_2 \in (0, 0.5] \) are randomly selected mixing coefficients, and \( \lambda_1, \lambda_2 \leq 0.5 \) ensures that the contribution of positive samples is always smaller than that of negative samples. \( k \) represents the index of any instance in the same cluster of positive samples to which sample \( i \) belongs in the same batch, and the specific generation rule for the same-cluster positive sample index set can be seen in Equation 16. \( j \) represents the index of any other sample in the same batch, excluding \( i \) and the indices in the same-cluster positive sample index set to which \( i \) belongs. 
When sample \( i \) does not have any positive samples from the same cluster, sample \( i \) itself is considered a positive sample.

As shown in Figure 3(b), the Clustering Universum hard negative samples (blue triangles) are close to the positive sample region in the embedding space but still maintain a certain distinction, making them suitable as high-quality hard negative samples. 
Finally, we adopt a simple and direct method to inject Clustering Universum hard negative samples into contrastive learning as additional hard negative samples in both time loss and instance loss (see Equations 3 and 5).

\subsubsection{Contrastive Loss}
\label{subsubsec:contrastive_loss}
This part introduces the contrastive loss function designed in our method. Traditional contrastive losses, such as instance-level contrastive loss, focus on maximizing the similarity between different augmentations of the same instance (positive sample pairs) while minimizing the similarity between augmentations of different instances (negative sample pairs)\cite{zhong2021graph}\cite{deng2023strongly}. However, this conflicts with the clustering objective, which aims to group samples from the same cluster together. Different from these traditional contrastive losses, we propose a cluster-aware contrastive loss function that incorporates both temporal and cluster-awareness information to guide the construction of positive and negative pairs, enabling the learning of cluster-friendly representations.

To achieve this, we incorporate two types of contrastive losses: time-level contrastive loss and instance-level contrastive loss.

For the time-level contrastive loss, we construct two augmented view pairs for each input: one pair combines cropping-based and perturbation-based views (\( \mathbf{z}^{(a)} \) and \( \mathbf{z}^{(b)} \)), while the other pair consists of two cropping-based views (\( \mathbf{z}^{(b)} \) and \( \mathbf{z}^{(c)} \)). 
Let \( i \) denote the index of the input time series, and \( t \) denote the corresponding timestamp. Thus, the overall time-level contrastive loss is defined as:  

\begin{equation}
    % \label{eq_1}
    L_{i,t}^{\text{temp}} = \hat{L}_{(i,t)}^{(a,b)} + \hat{L}_{(i,t)}^{(b,c)} 
\end{equation}

For the calculation of each component, we treat representations of the same timestamp as positive pairs and representations of different timestamps or time-level hard negatives as negative pairs. Without loss of generality, the time-level contrastive loss between the augmented views \( \mathbf{z}^{(a)}_{i,t} \) and \( \mathbf{z}^{(b)}_{i,t} \) of the \( i \)-th time series at timestamp \( t \) can be expressed as:

\begin{equation}
    % \label{eq_1}
    \resizebox{\columnwidth}{!}{%
    $\hat{L}_{i,t}^{(a,b)} = -\log \left( \frac{\exp \left( z_{i,t}^{(a)} \cdot z_{i,t}^{(b)} \right)}{\sum_{t' \in \Omega} \left( \exp \left( z_{i,t}^{(a)} \cdot z_{i,t'}^{(b)} \right) + \mathbb{I}_{(t \neq t')} \cdot \exp \left( z_{i,t}^{(a)} \cdot z_{i,t'}^{(a)} \right) + \exp \left( z_{i,t}^{(a)} \cdot \hat{h}_{i,t'}^{(a)} \right) + \exp \left( z_{i,t}^{(a)} \cdot \hat{h}_{i,t'}^{(b)} \right) \right)} \right)$}
\end{equation} 
where  \( \mathbb{I} \) is the indicator function, and \(\Omega\) represents the overlapping portion of the two time series crops. Additionally, \(z^{(a)}_{i,t}\) and \(z^{(b)}_{i,t}\) represent the \(a\) and \(b\) augmented representations of \(x_i\) at the same timestamp \(t\), while \(\hat{h}^{(a)}_{i,t'}\) and \(\hat{h}^{(b)}_{i,t'}\) represent the time-level hard negative samples generated from \(x_i\) at timestamp \(t'\) under the \(a\) and \(b\) augmentations. 

Similar to the time-level contrastive loss, the instance-level contrastive loss is also constructed from two augmented view pairs. The overall instance-level contrastive loss is defined as:

\begin{equation}
    % \label{eq_1}
    L_{i,t}^{\text{inst}} = \hat{L}_{(i,t)}^{(a,b)} + \hat{L}_{(i,t)}^{(b,c)}
\end{equation}

Compared with time-level contrastive loss, instance-level contrastive loss plays a more critical role in clustering tasks. In this loss calculation, we treat instances from the same cluster as positive samples and instances from different clusters or instance-level hard negatives as negative samples. By pulling same-cluster instances closer and pushing different-cluster instances apart, the model learns higher-level feature structures, effectively resolving the conflict between contrastive learning and clustering objectives. Without loss of generality, the instance-level contrastive loss between the augmented views \( \mathbf{z}^{(a)}_{i,t} \) and \( \mathbf{z}^{(b)}_{i,t} \) of the \( i \)-th time series at timestamp \( t \) can be expressed as:  

\begin{equation}
    % \label{eq_1}
    \resizebox{\columnwidth}{!}{
    $\hat{L}_{i,t}^{(a,b)} = -\log \left( \frac{\sum_{k \in N^+} \exp \left( z_{i,t}^{(a)} \cdot z_{k,t}^{(b)} \right)}{\sum_{j \in N^-} \left( \mathbb{I}_{(i \neq j)} \cdot \exp \left( z_{i,t}^{(a)} \cdot z_{j,t}^{(a)} \right) + \exp \left( z_{i,t}^{(a)} \cdot z_{j,t}^{(b)} \right) + \exp \left( z_{i,t}^{(a)} \cdot \hat{h}_{j,t}^{(a)} \right) + \exp \left( z_{i,t}^{(a)} \cdot \hat{h}_{j,t}^{(b)} \right) \right)} \right)$
}
\end{equation}  
where \( \hat{\mathbf{h}}^{(a)}_{j,t} \) and \( \hat{\mathbf{h}}^{(b)}_{j,t} \) represent the instance-level hard negative samples generated from \( \mathbf{x}_i \) under the \( a \) and \( b \) augmentations, respectively. Here, \( N^+ \) and \( N^- \) denote the sets of positive and negative sample indices. \( N^+ \) includes the index of sample \( i \) and its same-cluster positive samples within the same batch, while \( N^- \) includes the indices of other instances in the batch, excluding the same-cluster positive samples of \( i \). The specific generation rule for the same-cluster positive sample index set can be seen in Equation 16.

It is worth noting that if no same-cluster positive samples exist for \( \mathbf{x}_i \) in the batch, the loss function degenerates to the instance-level contrastive loss using regular mixed universe hard negative samples. This design not only captures instance-level feature relationships but also leverages clustering structures to enhance global cluster understanding, improving both feature discriminability and clustering performance.  

The total contrastive loss is the sum of the time-level contrastive loss and the instance-level contrastive loss. These two losses complement each other, capturing both the features of specific instances and the temporal variations. 
The total contrastive loss is defined as follows:
\begin{equation}
    % \label{eq_1}
    L_{\text{contrast}} = \frac{1}{NT} \sum_{i=1}^{N} \sum_{t=1}^{T} \left( L_{i,t}^{\text{temp}} + L_{i,t}^{\text{inst}} \right)
\end{equation}

\subsection{Fuzzy Clustering Module}
\label{subsec:fuccyclustering}
In real-world scenarios, the complex structure of time series data often leads to poorly separable clusters. However, most existing methods rely on hard clustering algorithms, assigning each instance to a single cluster with a fixed membership. These methods often perform poorly on complex time series data, particularly when handling overlapping clusters. To address this, we employ the fuzzy c-means\cite{bezdek1984fcm} (FCM) algorithm, which generates soft labels reflecting the degree of membership, allowing samples to belong to multiple clusters with varying degrees of membership. This flexible mechanism not only facilitates the handling of complex relationships in time series data but also helps select high-confidence sample relationships, providing more reliable guidance signals for contrastive learning and preventing interference from unreasonable relationships.  

The clustering loss function is defined as follows\cite{kim2004fuzzy}:  
\begin{equation}
    % \label{eq_1}
    L_{\text{cluster}} = \sum_{j=1}^{K} \sum_{i=1}^{N} p_{ij} \| \boldsymbol{z}_i - \boldsymbol{\mu}_j \|_2^2
\end{equation}
where \( p_{ij} \) represents the degree of membership of sample \( i \) to cluster \( j \), \( \boldsymbol{z}_i \) denotes the low-dimensional feature of sample \( i \), and \( \boldsymbol{\mu}_j \) is the center vector of cluster \( j \). Here, \( N \) is the number of data samples, and \( K \) is the number of clusters.  

Fuzzy clustering iteratively updates the cluster centers \( \boldsymbol{\mu}_j \) and membership degrees \( p_{ij} \). During each iteration, the membership degrees \( p_{ij} \) and cluster centers \( \boldsymbol{\mu}_j \) are updated according to the following formulas:  
\begin{equation}
    p_{ij} = \frac{\| \boldsymbol{z}_i - \boldsymbol{\mu}_j \|_2^{\frac{2}{1-m}}}{\sum_{j=1}^{K} \| \boldsymbol{z}_i - \boldsymbol{\mu}_j \|_2^{\frac{2}{1-m}}}
\end{equation}
\begin{equation}
    \boldsymbol{\mu}_j = \frac{\sum_{\boldsymbol{z}_i \in R} p_{ij}^m \cdot \boldsymbol{z}_i}{\sum_{r_i \in R} p_{ij}^m}
\end{equation}
where \( m \) represents the fuzziness degree, a weighting index that controls the overlap between clusters. Following the widely accepted standard proposed in \cite{sadeghi2021deep}, we fix \( m = 1.5 \) in this study.  

Finally, the total loss combines the contrastive loss and clustering loss, defined as:  
\begin{equation}
L = L_{\text{contrast}} + \alpha L_{\text{cluster}}
\end{equation}
where \( \alpha \) is a hyperparameter that balances the two losses.  

\subsection{Cluster-Awareness Generation Module}
To better jointly optimize representation learning and clustering, we design a cluster-awareness generation module that uses cluster structure information to guide the generation of index sets of positive samples in the same cluster, thereby providing more cluster-friendly representations for subsequent iterative optimization. To ensure both quantity control and quality assurance, we adopt a dual strategy when generating these index sets. 

Firstly, to extract the core cluster structure, we define the maximum degree of membership of each sample to its cluster center, along with the corresponding predicted cluster category, as follows:
\begin{equation}
    P_i = \max_{j=1, \dots, K} p_{ij}
\end{equation}
\begin{equation}
    c_i = \arg\max_{j=1, \dots, K} p_{ij}
\end{equation}

Next, we calculate the maximum number \( n \) of same-cluster positive samples to be selected from each cluster:

\begin{equation}
n = \left\lfloor r \cdot \frac{N}{K} \right\rfloor
\end{equation}
where \( r \) is the extraction ratio, \( N \) is the total number of samples, and \( K \) is the number of clusters. The samples in cluster \( j \) are sorted by their maximum membership values \( \{P_i \mid c_i = j\} \) in descending order, and the \( n \)-th largest membership value is selected as the quantity control threshold \( \xi_k \).

To further refine the selection, we define a quality control threshold \( \sigma_k \), which ensures that only samples with sufficiently high membership are selected. The threshold \( \sigma_k \) combines both quantity control (using \( \xi_k \)) and quality assurance (using the membership degree threshold \( \theta \)):

\begin{equation}
\sigma_k = \max(\xi_k, \theta)
\end{equation}

 Finally, the same-cluster positive sample index set \( C_k \) is defined as:  

\begin{equation}
C_k = \{i \mid P_i \geq \sigma_k, c_i = k\}
\end{equation}

\subsection{Implementation Details}
To mitigate the negative impact of poor initial cluster centers on subsequent optimization, we introduce a pre-training phase before the joint optimization stage. The membership information required by the cluster awareness generation module (the core component of the joint optimization) depends to a large extent on the initial cluster centers obtained by the FCM, while the initial cluster centers are influenced by the quality of the representation space. If the initial representation space is poorly structured, the calculated cluster centers may deviate significantly from the true data distribution, leading to suboptimal cluster structures that hinder further optimization.

To address this issue, the pre-training phase focuses on optimizing the initial representation space using a contrastive learning approach without the cluster-awareness generation module. Specifically, the pre-training process includes: (1) augmented view generation for time series data, (2) hard negative sample generation without cluster awareness, and (3) computation of degraded contrast loss (i.e., time-level contrast loss and instance-level contrast loss without cluster awareness). This ensures that the model learns to produce well-structured feature representations that are more discriminative and separable.

The joint optimization phase builds upon the pre-training phase by integrating the functions of all modules. During this phase, cluster-awareness information is dynamically adjusted, enabling the collaborative optimization of representation learning and clustering objectives. This joint optimization further refines the cluster structure, leading to improved clustering performance.

In the subsequent experimental section, we conducted detailed experiments to verify the role of the pre-training phase and the joint optimization phase. We validated the effectiveness of the pre-training phase in providing a more stable initialization of the cluster structure, as well as the importance of the joint optimization phase in achieving the collaborative optimization of representation learning and clustering objectives.
\section{Experimental Results}
\label{sec:experiments}

\subsection{Experimental Setup}
\label{subsec:setup}
\subsubsection{Dataset and Evaluation Metrics}
To validate the effectiveness of the proposed model, experiments are conducted on 40 datasets from the UCR time series dataset\cite{dau2019ucr}. Clustering performance is evaluated using two widely used metrics: Normalized Mutual Information\cite{strehl2002cluster} (NMI) and Rand Index\cite{rand1971objective} (RI). These metrics are extensively applied in the field of time series clustering and can comprehensively reflect the clustering performance of the model.

\subsubsection{Baseline Methods}
To comprehensively evaluate the performance of FCACC, this paper selects 8 representative algorithms as baseline methods for comparative experiments. These methods cover semi-supervised, self-supervised, and unsupervised representation learning, and include separated optimization methods and joint optimization methods. They are as follows:

DTC\cite{madiraju2018deep}: A deep time series clustering method that combines a time autoencoder with a clustering layer, learning the nonlinear clustering representation of time series by minimizing the KL divergence between the predicted and target distributions.

STCN\cite{ma2020self}: A self-supervised time series clustering network that integrates self-supervised learning with clustering tasks, ensuring both representation effectiveness and clustering accuracy.

TCGAN\cite{huang2023tcgan}: A generative adversarial framework for time series that uses adversarial training between the generator and discriminator to learn representations of time series data, thereby enhancing clustering performance.

FeatTS\cite{tiano2021featts}: A semi-supervised time series clustering method that introduces label information to guide feature selection, ensuring the selection of the most discriminative features for clustering.

TS2Vec\cite{yue2022ts2vec}: An unsupervised model for general time series representation learning, which employs hierarchical time contrastive learning to extract multi-scale features, excelling in tasks such as regression, classification, and prediction. For clustering tasks, FCM is applied.

TimesURL\cite{liu2024timesurl}: A time series representation learning framework that incorporates time-domain and frequency-domain enhancement strategies. It introduces a hard negative sample generation mechanism and achieves superior performance in downstream tasks such as prediction, classification, and anomaly detection by optimizing contrastive learning and reconstruction. We apply FCM for clustering tasks.

CDCC\cite{peng2024cross}: A time series clustering framework that integrates contrastive learning with both time and frequency domain information. It uses a dual-view autoencoder for representation learning and merges instance-level and cluster-level contrastive learning to enhance clustering-friendly optimization.

R-Clustering\cite{jorge2024time}: An efficient time series clustering method that utilizes random convolution kernels and principal component analysis (PCA). It combines random feature extraction and dimensionality reduction with K-means, achieving superior clustering performance and scalability.

\subsubsection{Model Architecture and Experiment Details}
For the baseline methods, we implement them using open-source code, following the parameter configurations and experimental setups specified in their respective papers. In the FCACC model, the specific parameter settings are \( \alpha = 0.2 \), \( r = 0.5 \), \( \theta = 0.95 \), and \( \lambda_1 = \lambda_2 = 0.2 \). The learning rate, batch size, and dropout rate are tuned through a search. To optimize the network training process, we choose the AdamW optimizer \cite{yao2021adahessian}. The experiments are conducted on a computer equipped with an AMD EPYC 9754 processor, a 4090D (24GB) GPU, and 60 GB of memory. The source code is publicly available\footnote{https://github.com/Du-Team/FCACC}.

The encoder $f$ consists of an input projection layer and an expanded CNN module. For each input $x_i$, the input projection layer is a fully connected layer that maps the observation $x_{i,t}$ at each timestamp $t$ into a high-dimensional space, forming a vector $z_{i,t}$. The expanded CNN module contains ten dilation blocks with residual blocks and one dilation block with a one-dimensional convolution layer. Each block includes two one-dimensional convolution layers with dilation parameters $(2^l$ for the $l$-th block), providing a larger receptive field\cite{bai2018empirical}.
\subsection{Overall Performance Comparing}
\begin{table*}[ht]
\caption{NMI of Different Methods on 40 Datasets.} % 添加标题
\begin{tabular}{c|ccccccccclllllll}
\cline{1-10}
\multirow{2}{*}{Dataset} &
  \multicolumn{9}{c}{\textbf{NMI}} &
   &
   &
   &
   &
   &
   &
   \\ \cline{2-10}
 &
  \textbf{DTC} &
  \textbf{STCN} &
  \textbf{TCGAN} &
  \textbf{FeatTS} &
  \textbf{TS2Vec} &
  \textbf{TimesURL} &
  \textbf{CDCC} &
  \textbf{R\_cluster} &
  \textbf{FCACC} &
   &
   &
   &
   &
   &
   &
   \\ \cline{1-10}
ACSF1 &
  0.348 &
  0.397 &
  0.208 &
  0.462 &
  \underline{0.551} &
  0.515 &
  0.35 &
  0.548 &
  \textbf{0.566} &
   &
   &
   &
   &
   &
   &
   \\
AllGestureWiimoteX &
  0.129 &
  0.152 &
  0.273 &
  \underline{0.305} &
  0.3 &
  0.263 &
  0 &
  0.28 &
  \textbf{0.329} &
   &
   &
   &
   &
   &
   &
   \\
AllGestureWiimoteZ &
  0.103 &
  0.133 &
  0.205 &
  0.191 &
  0.173 &
  \underline{0.257} &
  0 &
  0.228 &
  \textbf{0.332} &
   &
   &
   &
   &
   &
   &
   \\
BirdChicken &
  0 &
  0.029 &
  0.002 &
  0.399 &
  0.346 &
  0.029 &
  0.035 &
  \underline{0.421} &
  \textbf{0.464} &
   &
   &
   &
   &
   &
   &
   \\
Car &
  0.264 &
  0.269 &
  0.245 &
  0.24 &
  \textbf{0.526} &
  0.167 &
  0.26 &
  0.503 &
  \underline{0.51} &
   &
   &
   &
   &
   &
   &
   \\
CricketX &
  0.308 &
  0.174 &
  \underline{0.343} &
  0.258 &
  0.052 &
  0.244 &
  0.324 &
  0.337 &
  \textbf{0.372} &
   &
   &
   &
   &
   &
   &
   \\
CricketZ &
  0.276 &
  0.168 &
  \underline{0.337} &
  0.287 &
  0.059 &
  0.209 &
  0.28 &
  0.331 &
  \textbf{0.383} &
   &
   &
   &
   &
   &
   &
   \\
DistalPhalanxTW &
  0 &
  0.522 &
  0.555 &
  0.521 &
  0.467 &
  0.507 &
  0.505 &
  \underline{0.564} &
  \textbf{0.575} &
   &
   &
   &
   &
   &
   &
   \\
ECGFiveDays &
  0.01 &
  0.233 &
  0.002 &
  0.157 &
  0.24 &
  0.071 &
  \underline{0.526} &
  0.02 &
  \textbf{0.751} &
   &
   &
   &
   &
   &
   &
   \\
FaceAll &
  0.351 &
  0.313 &
  0.462 &
  0.288 &
  0.382 &
  0.345 &
  0.624 &
  \underline{0.661} &
  \textbf{0.721} &
   &
   &
   &
   &
   &
   &
   \\
FacesUCR &
  0.284 &
  0.358 &
  \underline{0.708} &
  0.321 &
  0.378 &
  0.268 &
  0.626 &
  0.648 &
  \textbf{0.756} &
   &
   &
   &
   &
   &
   &
   \\
Fungi &
  0.862 &
  0.738 &
  0.915 &
  0.717 &
  0.965 &
  0.935 &
  0.943 &
  \textbf{1} &
  \underline{0.979} &
   &
   &
   &
   &
   &
   &
   \\
GesturePebbleZ1 &
  0.11 &
  0.21 &
  0.382 &
  0.458 &
  \underline{0.459} &
  0.05 &
  0 &
  0.452 &
  \textbf{0.563} &
   &
   &
   &
   &
   &
   &
   \\
GesturePebbleZ2 &
  0.346 &
  0.165 &
  0.357 &
  0.328 &
  0.305 &
  0.055 &
  0 &
  \underline{0.455} &
  \textbf{0.542} &
   &
   &
   &
   &
   &
   &
   \\
GunPointOldVersusYoung &
  0.261 &
  0.979 &
  0.344 &
  0.949 &
  \textbf{1} &
  0.344 &
  \textbf{1} &
  \textbf{1} &
  \textbf{1} &
   &
   &
   &
   &
   &
   &
   \\
HouseTwenty &
  0.03 &
  0.096 &
  0.173 &
  \underline{0.578} &
  0.083 &
  0.005 &
  0.568 &
  0.247 &
  \textbf{0.71} &
   &
   &
   &
   &
   &
   &
   \\
InsectEPGRegularTrain &
  0.192 &
  0.164 &
  0.691 &
  0.606 &
  0.899 &
  \textbf{1} &
  0.616 &
  0.51 &
  \textbf{1} &
   &
   &
   &
   &
   &
   &
   \\
InsectEPGSmallTrain &
  0.197 &
  0.196 &
  \textbf{1} &
  0.642 &
  0.899 &
  \textbf{1} &
  0.615 &
  0.541 &
  \textbf{1} &
   &
   &
   &
   &
   &
   &
   \\
ItalyPowerDemand &
  0 &
  0.003 &
  0.198 &
  0.101 &
  \underline{0.282} &
  0 &
  0.089 &
  0 &
  \textbf{0.325} &
   &
   &
   &
   &
   &
   &
   \\
LargeKitchenAppliances &
  0.036 &
  \underline{0.172} &
  0.043 &
  0.113 &
  0.138 &
  0.022 &
  0.129 &
  0.005 &
  \textbf{0.266} &
   &
   &
   &
   &
   &
   &
   \\
Lightning2 &
  0.014 &
  0.021 &
  0.113 &
  \underline{0.138} &
  0.055 &
  0.134 &
  0.003 &
  0.03 &
  \textbf{0.201} &
   &
   &
   &
   &
   &
   &
   \\
Lightning7 &
  0.42 &
  0.302 &
  0.488 &
  0.364 &
  0.449 &
  0.403 &
  0.475 &
  \underline{0.53} &
  \textbf{0.562} &
   &
   &
   &
   &
   &
   &
   \\
Mallat &
  0.535 &
  0.652 &
  0.726 &
  0.587 &
  0.505 &
  \underline{0.905} &
  0.006 &
  0.887 &
  \textbf{0.944} &
   &
   &
   &
   &
   &
   &
   \\
Meat &
  0 &
  0.441 &
  0.521 &
  0.474 &
  0.576 &
  0.586 &
  0.014 &
  \textbf{0.708} &
  \underline{0.649} &
   &
   &
   &
   &
   &
   &
   \\
MedicalImages &
  0.231 &
  0.249 &
  0.246 &
  0.188 &
  0.165 &
  0.176 &
  0.258 &
  \textbf{0.31} &
  \underline{0.295} &
   &
   &
   &
   &
   &
   &
   \\
MiddlePhalanxOutlineAgeGroup &
  0.03 &
  0.394 &
  0.397 &
  0.227 &
  0.391 &
  0.368 &
  0.391 &
  \textbf{0.4} &
  \textbf{0.4} &
   &
   &
   &
   &
   &
   &
   \\
PickupGestureWiimoteZ &
  0.366 &
  0.548 &
  \underline{0.719} &
  0.444 &
  0.705 &
  0.432 &
  0 &
  0.602 &
  \textbf{0.749} &
   &
   &
   &
   &
   &
   &
   \\
PigAirwayPressure &
  0.614 &
  0.438 &
  0.562 &
  0.551 &
  \underline{0.834} &
  0.498 &
  0.537 &
  0.632 &
  \textbf{0.883} &
   &
   &
   &
   &
   &
   &
   \\
PigCVP &
  0.585 &
  0.486 &
  0.527 &
  0.64 &
  \underline{0.836} &
  0.754 &
  0.592 &
  0.7 &
  \textbf{0.882} &
   &
   &
   &
   &
   &
   &
   \\
Plane &
  0.339 &
  0.936 &
  0.932 &
  0.683 &
  \textbf{0.982} &
  0.971 &
  0.961 &
  \textbf{0.982} &
  \textbf{0.982} &
   &
   &
   &
   &
   &
   &
   \\
ProximalPhalanxOutlineAgeGroup &
  0.522 &
  0.496 &
  0.534 &
  0.504 &
  0.497 &
  0.428 &
  0.495 &
  \textbf{0.566} &
  \underline{0.549} &
   &
   &
   &
   &
   &
   &
   \\
SemgHandMovementCh2 &
  0.119 &
  0.112 &
  0.231 &
  0.188 &
  0.232 &
  0.233 &
  0.246 &
  \underline{0.25} &
  \textbf{0.267} &
   &
   &
   &
   &
   &
   &
   \\
ShapeletSim &
  0.004 &
  0.605 &
  0.015 &
  0.806 &
  \textbf{1} &
  0.033 &
  0.32 &
  \textbf{1} &
  \textbf{1} &
   &
   &
   &
   &
   &
   &
   \\
ShapesAll &
  0.65 &
  0.485 &
  0.714 &
  0.512 &
  0.469 &
  0.387 &
  0.635 &
  \textbf{0.756} &
  \underline{0.737} &
   &
   &
   &
   &
   &
   &
   \\
SmoothSubspace &
  0.313 &
  0.316 &
  0.369 &
  0.026 &
  \textbf{0.537} &
  0.234 &
  0.394 &
  0.304 &
  \underline{0.438} &
   &
   &
   &
   &
   &
   &
   \\
Symbols &
  0.667 &
  0.72 &
  0.84 &
  0.699 &
  0.92 &
  0.724 &
  0.82 &
  \textbf{0.952} &
  \underline{0.924} &
   &
   &
   &
   &
   &
   &
   \\
SyntheticControl &
  0.671 &
  0.582 &
  \underline{0.808} &
  0.482 &
  0.791 &
  0.805 &
  0.792 &
  0.806 &
  \textbf{0.984} &
   &
   &
   &
   &
   &
   &
   \\
ToeSegmentation1 &
  0.001 &
  0.017 &
  0 &
  0.166 &
  \underline{0.423} &
  0.033 &
  0.069 &
  0.027 &
  \textbf{0.468} &
   &
   &
   &
   &
   &
   &
   \\
ToeSegmentation2 &
  0.025 &
  0.037 &
  0.026 &
  0.105 &
  \underline{0.284} &
  0.003 &
  0.258 &
  0.205 &
  \textbf{0.398} &
   &
   &
   &
   &
   &
   &
   \\
TwoPatterns &
  0.016 &
  0.116 &
  0.005 &
  0.3 &
  0.208 &
  0.006 &
  0.018 &
  \underline{0.32} &
  \textbf{0.807} &
   &
   &
   &
   &
   &
   &
   \\ \cline{1-10}
\textbf{Average   NMI} &
  0.256 &
  0.336 &
  0.405 &
  0.4 &
  0.484 &
  0.36 &
  0.369 &
  0.493 &
  \textbf{0.632} &
   &
   &
   &
   &
   &
   &
   \\
\textbf{Best} &
  0 &
  0 &
  1 &
  0 &
  5 &
  2 &
  1 &
  10 &
  \textbf{32} &
   &
   &
   &
   &
   &
   &
   \\
\textbf{AVERAGE   RANK} &
  7.2 &
  6.425 &
  4.9 &
  5.625 &
  4.2 &
  6.1 &
  5.5 &
  3.275 &
  \textbf{1.2} &
   &
   &
   &
   &
   &
   &
   \\ \cline{1-10}
\end{tabular}
\end{table*}

% Please add the following required packages to your document preamble:
% \usepackage{multirow}
% \usepackage[normalem]{ulem}
% \useunder{\uline}{\ul}{}  \caption{RI of different methods on 40 datasets.} % 添加标题
% Please add the following required packages to your document preamble:
% \usepackage{multirow}
% \usepackage[normalem]{ulem}
% \useunder{\uline}{\ul}{}

\begin{table*}[ht]
\caption{RI of Different Methods on 40 Datasets.} % 添加标题
\begin{tabular}{c|cccccccccllllll}
\cline{1-10}
\multirow{2}{*}{Dataset} &
  \multicolumn{9}{c}{\textbf{RI}} &
   &
   &
   &
   &
   &
   \\ \cline{2-10}
 &
  \textbf{DTC} &
  \textbf{STCN} &
  \textbf{TCGAN} &
  \textbf{FeatTS} &
  \textbf{TS2Vec} &
  \textbf{TimesURL} &
  \textbf{CDCC} &
  \textbf{R\_cluster} &
  \textbf{FCACC} &
   &
   &
   &
   &
   &
   \\ \cline{1-10}
ACSF1 &
  0.689 &
  0.75 &
  0.294 &
  0.841 &
  \textbf{0.888} &
  0.796 &
  0.84 &
  0.85 &
  \underline{0.858} &
   &
   &
   &
   &
   &
   \\
AllGestureWiimoteX &
  0.791 &
  0.773 &
  0.724 &
  0.701 &
  0.762 &
  \underline{0.819} &
  0.099 &
  0.816 &
  \textbf{0.843} &
   &
   &
   &
   &
   &
   \\
AllGestureWiimoteZ &
  0.659 &
  0.805 &
  0.803 &
  0.787 &
  0.754 &
  \underline{0.824} &
  0.099 &
  0.82 &
  \textbf{0.849} &
   &
   &
   &
   &
   &
   \\
BirdChicken &
  0.474 &
  0.508 &
  0.488 &
  \textbf{0.738} &
  \underline{0.704} &
  0.508 &
  0.499 &
  0.672 &
  \underline{0.704} &
   &
   &
   &
   &
   &
   \\
Car &
  0.683 &
  0.705 &
  0.64 &
  0.682 &
  \underline{0.795} &
  0.659 &
  0.71 &
  0.721 &
  \textbf{0.799} &
   &
   &
   &
   &
   &
   \\
CricketX &
  0.855 &
  0.824 &
  0.849 &
  0.864 &
  0.519 &
  0.75 &
  \underline{0.873} &
  0.869 &
  \textbf{0.879} &
   &
   &
   &
   &
   &
   \\
CricketZ &
  0.845 &
  0.822 &
  0.852 &
  0.866 &
  0.52 &
  0.715 &
  \underline{0.868} &
  0.864 &
  \textbf{0.875} &
   &
   &
   &
   &
   &
   \\
DistalPhalanxTW &
  0.212 &
  0.807 &
  \underline{0.888} &
  0.824 &
  0.783 &
  0.791 &
  0.764 &
  0.808 &
  \textbf{0.9} &
   &
   &
   &
   &
   &
   \\
ECGFiveDays &
  0.505 &
  0.652 &
  0.5 &
  0.6 &
  0.648 &
  0.548 &
  \underline{0.817} &
  0.513 &
  \textbf{0.899} &
   &
   &
   &
   &
   &
   \\
FaceAll &
  0.853 &
  0.871 &
  0.893 &
  0.873 &
  0.77 &
  0.797 &
  \underline{0.925} &
  0.916 &
  \textbf{0.937} &
   &
   &
   &
   &
   &
   \\
FacesUCR &
  0.854 &
  0.873 &
  \underline{0.935} &
  0.837 &
  0.718 &
  0.676 &
  0.926 &
  0.919 &
  \textbf{0.946} &
   &
   &
   &
   &
   &
   \\
Fungi &
  0.962 &
  0.937 &
  0.974 &
  0.917 &
  0.991 &
  0.979 &
  0.984 &
  \textbf{1} &
  \underline{0.992} &
   &
   &
   &
   &
   &
   \\
GesturePebbleZ1 &
  0.248 &
  0.764 &
  0.779 &
  0.796 &
  \underline{0.803} &
  0.653 &
  0.164 &
  0.792 &
  \textbf{0.841} &
   &
   &
   &
   &
   &
   \\
GesturePebbleZ2 &
  0.597 &
  0.755 &
  0.782 &
  0.686 &
  0.73 &
  0.708 &
  0.164 &
  \underline{0.792} &
  \textbf{0.835} &
   &
   &
   &
   &
   &
   \\
GunPointOldVersusYoung &
  0.608 &
  0.995 &
  0.619 &
  0.987 &
  \textbf{1} &
  0.619 &
  \textbf{1} &
  \textbf{1} &
  \textbf{1} &
   &
   &
   &
   &
   &
   \\
HouseTwenty &
  0.522 &
  0.566 &
  0.586 &
  0.818 &
  0.553 &
  0.498 &
  \underline{0.838} &
  0.662 &
  \textbf{0.904} &
   &
   &
   &
   &
   &
   \\
InsectEPGRegularTrain &
  0.616 &
  0.586 &
  0.785 &
  0.855 &
  0.97 &
  \textbf{1} &
  0.793 &
  0.75 &
  \textbf{1} &
   &
   &
   &
   &
   &
   \\
InsectEPGSmallTrain &
  0.611 &
  0.603 &
  \textbf{1} &
  0.872 &
  0.968 &
  \textbf{1} &
  0.793 &
  0.776 &
  \textbf{1} &
   &
   &
   &
   &
   &
   \\
ItalyPowerDemand &
  0.5 &
  0.501 &
  0.552 &
  0.568 &
  \underline{0.677} &
  0.5 &
  0.56 &
  0.5 &
  \textbf{0.697} &
   &
   &
   &
   &
   &
   \\
LargeKitchenAppliances &
  0.561 &
  \underline{0.645} &
  0.575 &
  0.605 &
  0.627 &
  0.546 &
  0.622 &
  0.549 &
  \textbf{0.683} &
   &
   &
   &
   &
   &
   \\
Lightning2 &
  0.498 &
  0.514 &
  0.514 &
  \textbf{0.566} &
  0.521 &
  0.533 &
  0.499 &
  0.517 &
  \underline{0.554} &
   &
   &
   &
   &
   &
   \\
Lightning7 &
  0.794 &
  0.77 &
  0.771 &
  0.733 &
  0.81 &
  0.793 &
  \underline{0.828} &
  0.816 &
  \textbf{0.838} &
   &
   &
   &
   &
   &
   \\
Mallat &
  0.672 &
  0.895 &
  0.882 &
  0.863 &
  0.709 &
  0.955 &
  0.775 &
  \underline{0.964} &
  \textbf{0.985} &
   &
   &
   &
   &
   &
   \\
Meat &
  0.322 &
  0.704 &
  0.739 &
  0.766 &
  0.781 &
  \underline{0.789} &
  0.558 &
  \textbf{0.861} &
  0.721 &
   &
   &
   &
   &
   &
   \\
MedicalImages &
  0.662 &
  0.67 &
  0.66 &
  0.668 &
  0.643 &
  0.675 &
  \underline{0.682} &
  0.678 &
  \textbf{0.689} &
   &
   &
   &
   &
   &
   \\
MiddlePhalanxOutlineAgeGroup &
  0.5 &
  0.732 &
  0.732 &
  0.669 &
  0.734 &
  0.729 &
  \underline{0.735} &
  0.733 &
  \textbf{0.736} &
   &
   &
   &
   &
   &
   \\
PickupGestureWiimoteZ &
  0.635 &
  0.866 &
  \underline{0.914} &
  0.801 &
  0.9 &
  0.843 &
  0.091 &
  0.876 &
  \textbf{0.919} &
   &
   &
   &
   &
   &
   \\
PigAirwayPressure &
  0.958 &
  0.906 &
  0.963 &
  0.965 &
  \underline{0.972} &
  0.911 &
  0.961 &
  0.968 &
  \textbf{0.985} &
   &
   &
   &
   &
   &
   \\
PigCVP &
  0.953 &
  0.912 &
  0.879 &
  0.962 &
  \underline{0.978} &
  0.972 &
  0.963 &
  0.971 &
  \textbf{0.985} &
   &
   &
   &
   &
   &
   \\
Plane &
  0.375 &
  0.958 &
  0.954 &
  0.873 &
  0.995 &
  0.992 &
  0.99 &
  \textbf{0.995} &
  \textbf{0.995} &
   &
   &
   &
   &
   &
   \\
ProximalPhalanxOutlineAgeGroup &
  0.788 &
  0.784 &
  \underline{0.798} &
  0.775 &
  0.786 &
  0.69 &
  0.784 &
  0.796 &
  \textbf{0.804} &
   &
   &
   &
   &
   &
   \\
SemgHandMovementCh2 &
  0.718 &
  0.699 &
  0.62 &
  0.561 &
  0.757 &
  0.72 &
  \textbf{0.764} &
  0.741 &
  \textbf{0.764} &
   &
   &
   &
   &
   &
   \\
ShapeletSim &
  0.499 &
  0.827 &
  0.507 &
  0.942 &
  \textbf{1} &
  0.52 &
  0.703 &
  \textbf{1} &
  \textbf{1} &
   &
   &
   &
   &
   &
   \\
ShapesAll &
  0.968 &
  0.905 &
  0.964 &
  0.917 &
  0.848 &
  0.809 &
  0.968 &
  \textbf{0.982} &
  \underline{0.976} &
   &
   &
   &
   &
   &
   \\
SmoothSubspace &
  0.682 &
  0.694 &
  0.717 &
  0.562 &
  \underline{0.74} &
  0.65 &
  0.667 &
  0.663 &
  \textbf{0.752} &
   &
   &
   &
   &
   &
   \\
Symbols &
  0.791 &
  0.884 &
  0.917 &
  0.859 &
  0.975 &
  0.883 &
  0.934 &
  \textbf{0.987} &
  \underline{0.976} &
   &
   &
   &
   &
   &
   \\
SyntheticControl &
  0.84 &
  0.83 &
  0.872 &
  0.768 &
  0.908 &
  0.927 &
  \underline{0.922} &
  0.897 &
  \textbf{0.997} &
   &
   &
   &
   &
   &
   \\
ToeSegmentation1 &
  0.498 &
  0.51 &
  0.498 &
  0.609 &
  \underline{0.73} &
  0.522 &
  0.545 &
  0.517 &
  \textbf{0.778} &
   &
   &
   &
   &
   &
   \\
ToeSegmentation2 &
  0.5 &
  0.513 &
  0.504 &
  0.543 &
  0.602 &
  0.497 &
  0.602 &
  \underline{0.694} &
  \textbf{0.726} &
   &
   &
   &
   &
   &
   \\
TwoPatterns &
  0.623 &
  0.655 &
  0.622 &
  \underline{0.732} &
  0.686 &
  0.575 &
  0.631 &
  0.725 &
  \textbf{0.912} &
   &
   &
   &
   &
   &
   \\ \cline{1-10}
\textbf{Average   RI} &
  0.648 &
  0.749 &
  0.739 &
  0.771 &
  0.781 &
  0.734 &
  0.699 &
  0.799 &
  \textbf{0.863} &
   &
   &
   &
   &
   &
   \\
\textbf{Best} &
  0 &
  0 &
  1 &
  2 &
  3 &
  2 &
  2 &
  7 &
  \textbf{33} &
   &
   &
   &
   &
   &
   \\
\textbf{AVERAGE   RANK} &
  7.275 &
  5.95 &
  5.725 &
  5.5 &
  4.2 &
  5.9 &
  4.825 &
  3.625 &
  \textbf{1.275} &
   &
   &
   &
   &
   &
   \\ \cline{1-10}
\end{tabular}
\end{table*}

In Tables I and II, we compare the proposed FCACC method with DTC, STCN, TCGAN, FeatTS, TS2Vec, TimesURL, CDCC, and R-Clustering. Bold text indicates that the method ranks first on the corresponding dataset, while underlined text indicates that the method ranks second. FCACC achieves the best NMI in 32 out of 40 datasets and the best RI in 33 datasets. It also achieves the highest average NMI (0.632), the highest average RI (0.863), and the highest average rankings, with 1.2 (NMI) and 1.275 (RI), respectively. Notably, FCACC achieves an NMI and RI of 1 on the GunPointOldVersusYoung, ShapeletSim, InsectEPGRegularTrain, and InsectEPGSmallTrain datasets.

In the comparison of various deep clustering methods, FCACC performs excellently on the majority of datasets. Compared to methods based on hard clustering (e.g., DTC, R-Clustering), FCACC effectively addresses the ambiguity and uncertainty in time series data through soft clustering, which enables it to demonstrate greater adaptability in complex clustering tasks. Compared to separated optimization methods (e.g., TS2Vec), FCACC further enhances clustering performance by jointly optimizing representation learning and clustering objectives. In comparison to traditional joint optimization methods (e.g., STCN, TCGAN, etc.), FCACC establishes deeper collaborative optimization between the representation and clustering objectives, achieving higher-quality clustering. Although CDCC integrates multi-dimensional information, FCACC leverages a three-view data augmentation strategy based on multiple clipping and perturbations, utilizing various features of time series to achieve stronger clustering performance. Additionally, even without labels, FCACC demonstrates better clustering results than the semi-supervised clustering method FeatTS in certain aspects, further showcasing its potential.
\subsection{Ablation Study}
To evaluate the effectiveness of the components in FCACC, Table III presents a comparison between the complete FCACC and its four variants across 40 UCR datasets. Specifically: (1) \textit{w/o Three-View Data Augmentation} replaces the three-view data augmentation strategy with a context-based two-view augmentation strategy\cite{yue2022ts2vec}; (2) \textit{w/o Cluster-Aware Hard Negative Sample Generation} removes the proposed hard negative sample generation mechanism; (3) \textit{w/o Cluster-Aware Positive-Negative Sample Pair Selection} eliminates the cluster-aware guidance in selecting positive and negative samples for contrastive loss calculation; (4) \textit{w/o Cluster-Awareness Generation} substitutes the cluster-awareness generation module with clustering labels generated via $\arg\max$. The results demonstrate that each component in FCACC plays a crucial role in its overall performance.
% Please add the following required packages to your document preamble:
% \usepackage{booktabs}
\begin{table}[htbp]
\centering
\caption{Ablation Study}
\label{tab:ablation}
\begin{tabular}{@{}cc@{}} % Both columns are centered
\toprule
\multicolumn{1}{c}{Method} & \multicolumn{1}{c}{\textbf{Avg.NMI}} \\ % Title row centered
\midrule
\multicolumn{1}{c}{\textbf{FCACC}} & \multicolumn{1}{c}{\textbf{0.632}} \\ % FCACC row centered
w/o Three-View Data Augmentation      & 0.563 (-6.9\%)     \\
w/o Cluster-Aware Hard Negative Sample Generation   & 0.555 (-7.7\%)     \\
w/o Cluster-Aware Positive-Negative Sample Pair Selection    & 0.606 (-2.6\%)     \\
w/o Cluster-Awareness Generation     & 0.614 (-1.8\%)     \\ 
\bottomrule
\end{tabular}
\end{table}

The ablation experiment results demonstrate that FCACC's performance relies on the collaborative functioning of multiple modules. First, removing the three-view data augmentation strategy reduces the average NMI to 0.563 (a 6.9\% decrease), indicating that multi-view augmentation is essential for capturing the complex features of time series. Second, removing the cluster-aware hard negative sample generation strategy decreases the NMI by 7.7\%, highlighting its role in enhancing the model’s discriminative ability through more challenging negative samples. Furthermore, removing the cluster-aware positive-negative sample pair selection strategy results in a 2.6\% NMI drop, while removing the cluster-awareness generation module leads to a 1.8\% reduction, underscoring the importance of both modules in achieving joint optimization between representation learning and clustering tasks. Overall, the analysis confirms that each module of FCACC contributes significantly to its final performance.

\subsection{Effectiveness Analysis of Joint Optimization}
\begin{table}[ht]
\centering
\caption{Analysis of Effectiveness of Joint Optimization}
\begin{tabular}{cc}
\hline
                                 & \textbf{Avg.NMI} \\ \hline
\textbf{FCACC}                   & \textbf{0.632}   \\
Joint Optimization Only               & 0.505 (-12.7\%)  \\
Pre-trained Representation + FCM & 0.514 (-11.8\%)  \\ \hline
\end{tabular}
\end{table}
\subsubsection{Analysis of Impact of Joint Optimization and Pre-training on Clustering Performance}
In the experiments, we compare two variants across 40 UCR datasets: one trained using only the joint optimization phase, and the other trained using only the pre-training phase followed by clustering with the same fuzzy coefficient in FCM. 
We find that the clustering performance of both variants is significantly lower than that of the proposed method. 
The main reason for this is the failure to fully exploit the collaborative effect between representation learning and the clustering objective. 
For pure joint optimization, representation learning and clustering are performed simultaneously in the initial stage. Since the initial representations obtained by the model are poor, the generated initial cluster centers are of low quality. The cluster structures derived from these low-quality cluster centers guide representation learning, which can have a sustained negative impact on subsequent optimization and may be difficult to correct effectively during the iterative process, thereby affecting the final clustering performance. For the variant using only the pre-training phase, although preliminary feature representations are obtained, the absence of clustering structure involvement in the optimization means that the feature space lacks an understanding of the cluster structure, making it difficult to effectively complete the clustering task.

The experiments demonstrate the necessity of adding an additional pre-training phase and establishing an effective collaboration between representation learning and the clustering objective. The proposed FCACC method stabilizes the feature space initialization through the pre-training phase, providing a high-quality starting point for joint optimization. The joint optimization phase dynamically generates cluster structures with fuzzy membership, enabling collaborative optimization between representation learning and clustering, thereby significantly improving clustering performance.

\subsubsection{Analysis of Change in Number of Cluster-aware Samples}
Figure 4 illustrates the trend of cluster-aware sample quantity throughout the training process. As training progresses, the number of samples selected by the cluster-awareness mechanism gradually increases, reflecting the model’s improving ability to capture and express cluster structures. 

In the early stages of training, the number of cluster-aware samples remains relatively low due to the limited clustering quality, which has not yet been enhanced through joint optimization. Only a few samples meet the membership threshold \( \theta \) and are selected for the cluster-aware set. As joint optimization advances, clustering quality improves, and more samples meet the membership threshold \( \theta \) without exceeding the extraction ratio \( r \). Consequently, these samples are included in the cluster-aware set. 

This process creates a positive feedback loop: as more high-quality cluster-aware samples are included in the set, the model’s understanding and expression of the cluster structure strengthen. This, in turn, leads to better alignment between the clustering objective and the feature representation, further improving clustering performance.
% \begin{figure}[ht]
%     \centering
%     \includegraphics[width=1\linewidth]{figures/num_of_cluster.png}
%     \caption{Change in the number of cluster-aware samples during training, reflecting the model's progressively enhanced understanding of the cluster structure.}
%     \label{fig:enter-label}
% \end{figure}
\begin{figure}[!ht]
    \centering
    \subfigure[FacesUCR Cluster Number]{
        \centering
        \includegraphics[width=0.46\linewidth]{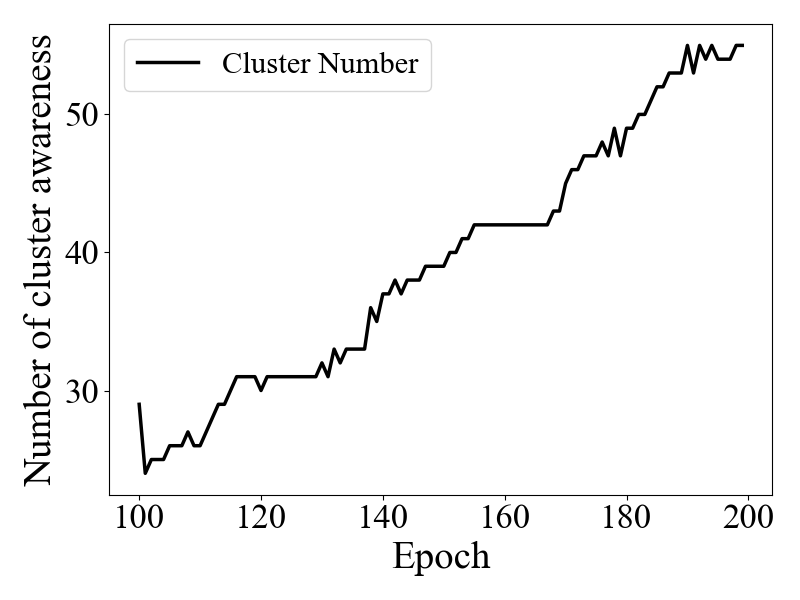}  % 设置子图宽度为33%
        \label{subfig:clu_c}   
    }
    \subfigure[\small ProximalPhalanxOutlineAgeGroup Cluster Number]{
        \centering
        \includegraphics[width=0.46\linewidth]{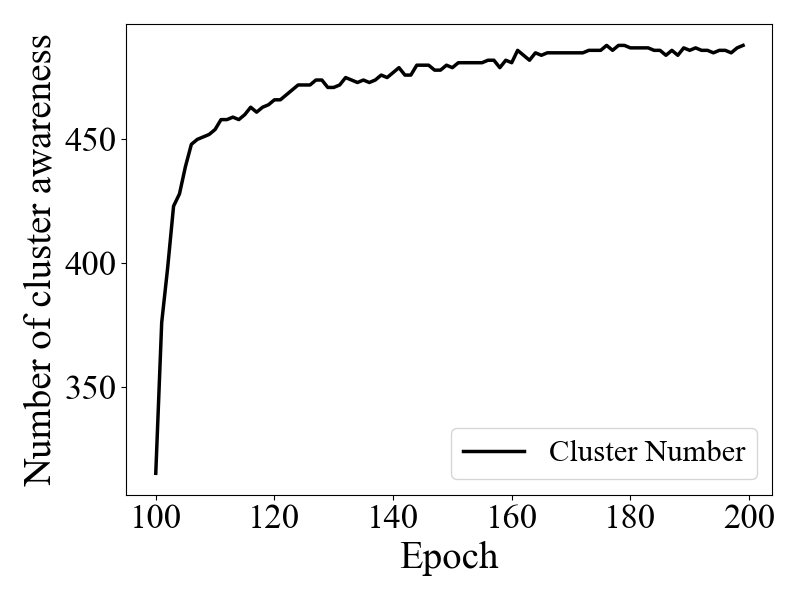}  % 设置子图宽度为33%
        \label{subfig:clu_d}
    }
    
    \caption{Variation in Cluster-Aware Samples During Training. The model's progressively enhanced understanding of the cluster structure is reflected in this change.
    }
    % \label{fig:example} %\ref{subfig:parameter_c}
\end{figure}

\subsection{Parameter Analysis}
% \begin{figure}[ht]
%     \centering
%     \includegraphics[width=1\linewidth]{figures/parameter_analysis.png}
%     \caption{Effect of Parameters.}
%     \label{fig:enter-label}
% \end{figure}
\begin{figure*}[!ht]
    \centering
    \subfigure[Ratio \( r \)]{
        \centering
        \includegraphics[width=0.3\linewidth]{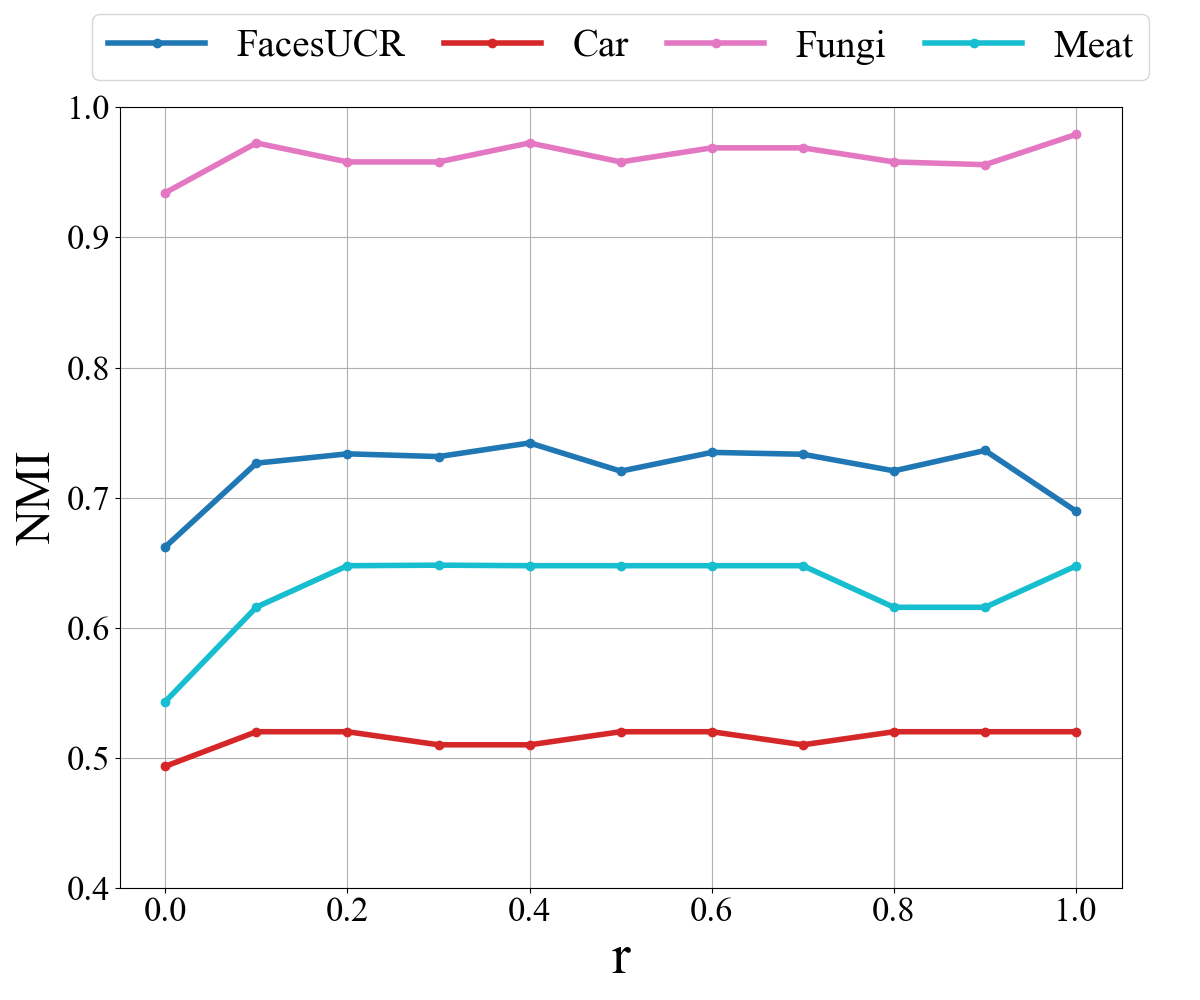}  % 设置子图宽度为33%
        \label{subfig:parameter_a}
    }
    % \hspace{0.01\linewidth}  % 控制子图之间的间距
    \subfigure[Threshold \( \theta \)]{
        \centering
        \includegraphics[width=0.3\linewidth]{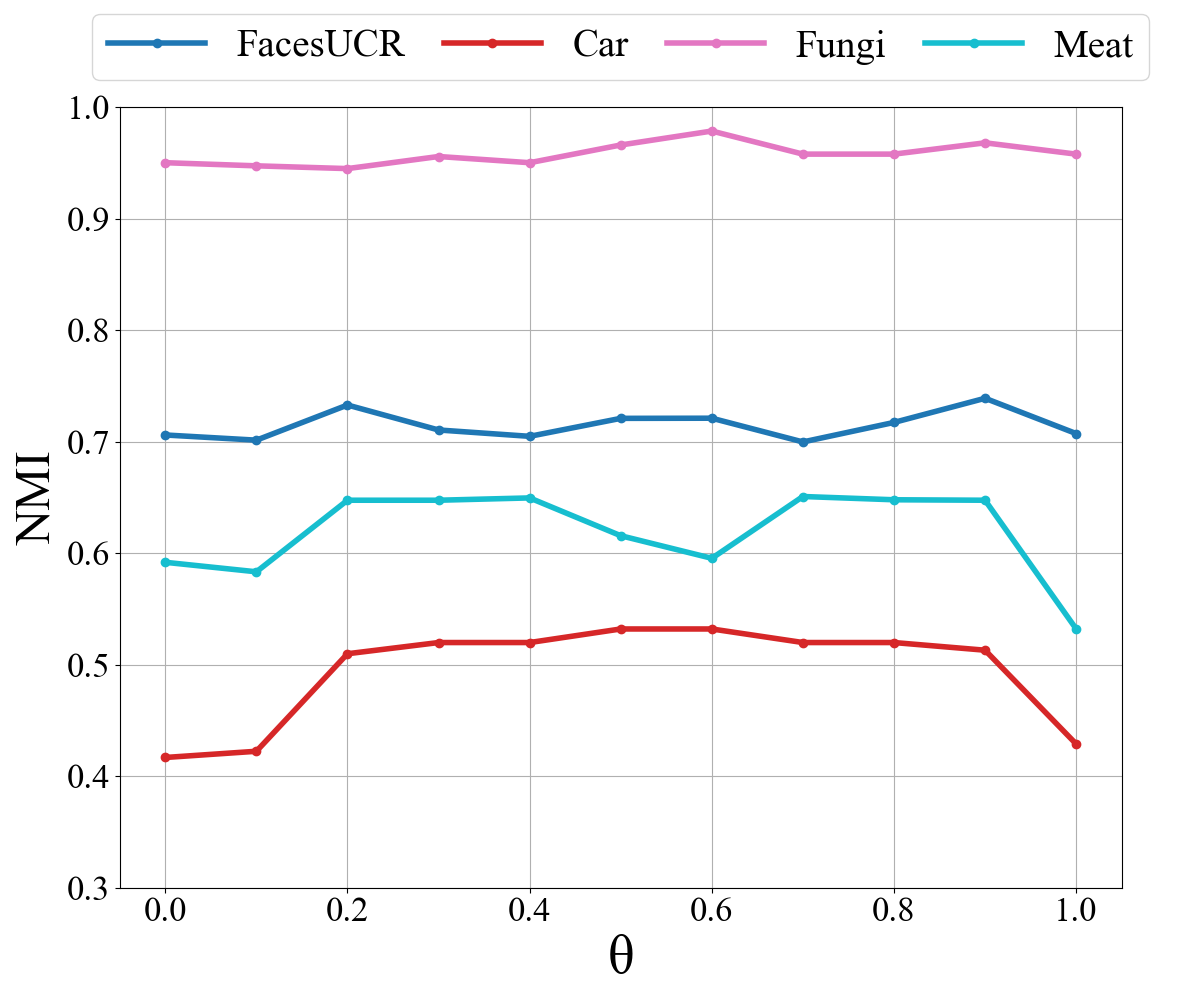}  % 设置子图宽度为33%
        \label{subfig:parameter_b}
    }
    % \hspace{0.01\linewidth}  % 控制子图之间的间距
    \subfigure[Loss Weight \( \alpha \)]{
        \centering
        \includegraphics[width=0.3\linewidth]{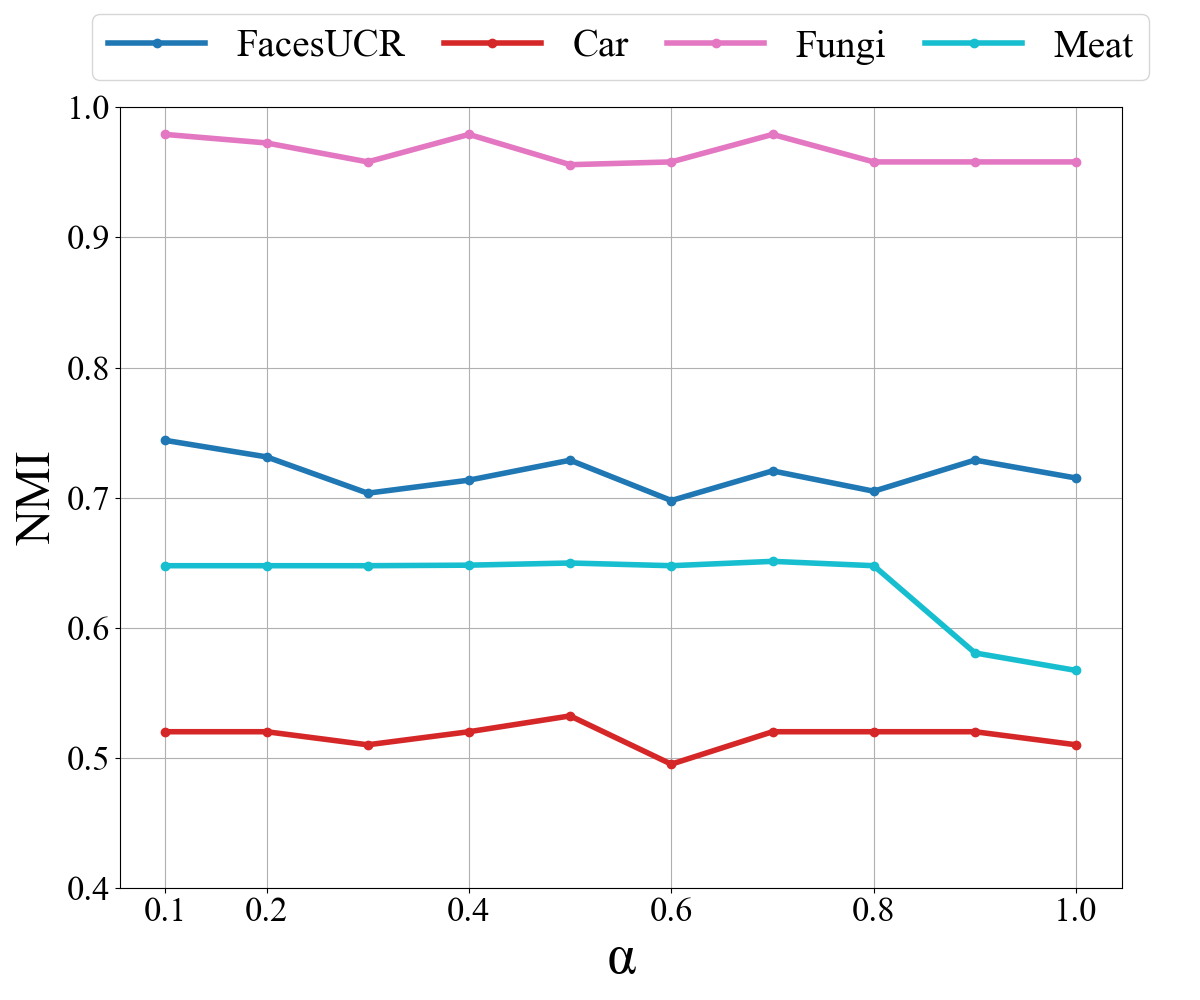}  % 设置子图宽度为33%
        \label{subfig:parameter_c}
    }
    \caption{
        Effect of Parameters.
    }
    % \label{fig:example} %\ref{subfig:parameter_c}
\end{figure*}

In this experiment, we performed a parameter sensitivity analysis on three key parameters of the FCACC model—cluster-aware extraction ratio (\( r \)), membership degree threshold (\( \theta \)), and clustering loss weight (\( \alpha \))—across four UCR datasets: FacesUCR, Car, Fungi, and Meat. These parameters control critical aspects of different mechanisms in the model.

The experimental results show that the impact of these parameters on model performance varies across different datasets, and they play an important role in optimizing performance.

The cluster-aware extraction ratio (\( r \)) controls the maximum number of samples extracted from each cluster, which is the core parameter determining the number of samples within a cluster. \( \theta \) is the cluster-aware quality filtering threshold, which ensures the quality of the cluster structure by removing samples with low membership degrees. As shown in Figures 5(a) and 5(b), performance remains stable within a reasonable range of \( r \) and \( \theta \). However, when one of these criteria is abandoned (i.e., \( r = 0 \) or \( \theta = 1.0 \)), performance declines to varying degrees. These results highlight the robustness and necessity of the dual criteria.

\( \alpha \) is the clustering loss balance coefficient, used to adjust the weight ratio between clustering loss and contrastive learning loss, affecting the balance between representation learning and clustering objectives. The impact of \( \alpha \) on model performance is shown in Figure 5(c). The experimental results indicate that within a reasonable range, different values of \( \alpha \) yield stable performance. This may be because our improvements allow the contrastive learning loss to also partially capture the cluster structure, thus avoiding excessive reliance on the \( \alpha \) value.
\subsection{Convergence Analysis}
In this subsection, we evaluate the convergence of the proposed FCACC method by reporting the changes in loss values during training and the corresponding clustering performance.
\subsubsection{Loss Variation During Training Process}
\begin{figure}[!ht]
    \centering
    \subfigure[Pretraining Loss of FacesUCR]{
        \centering
        \includegraphics[width=0.46\linewidth]{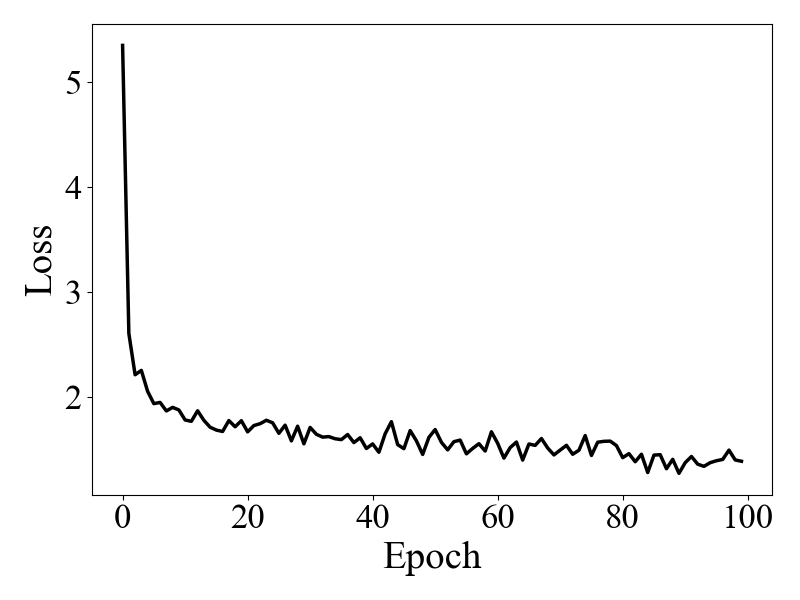}  % 设置子图宽度为33%
        % \label{subfig:loss_c}
    }
    \subfigure[Joint Optimization Loss of FacesUCR]{
        \centering
        \includegraphics[width=0.46\linewidth]{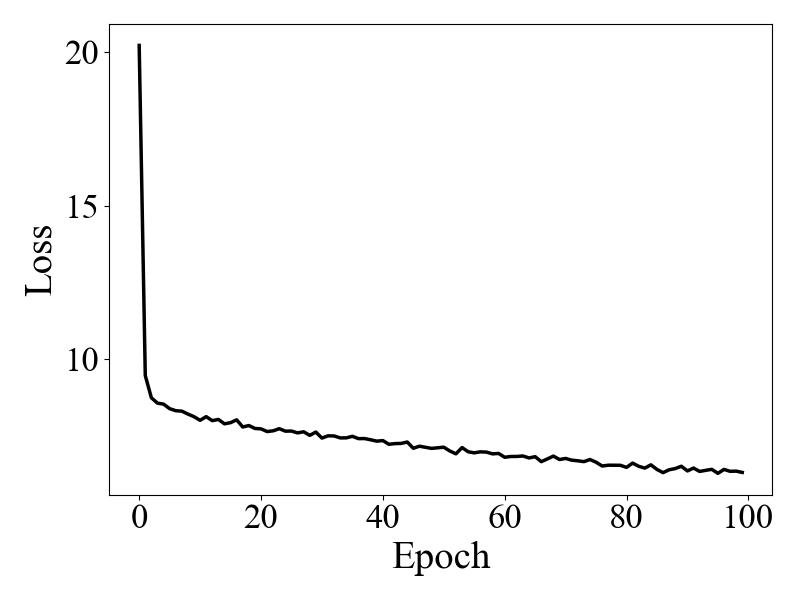}  % 设置子图宽度为33%
        % \label{subfig:loss_d}
    }
    \subfigure[Pretraining Loss of HouseTwenty]{
        \centering
        \includegraphics[width=0.46\linewidth]{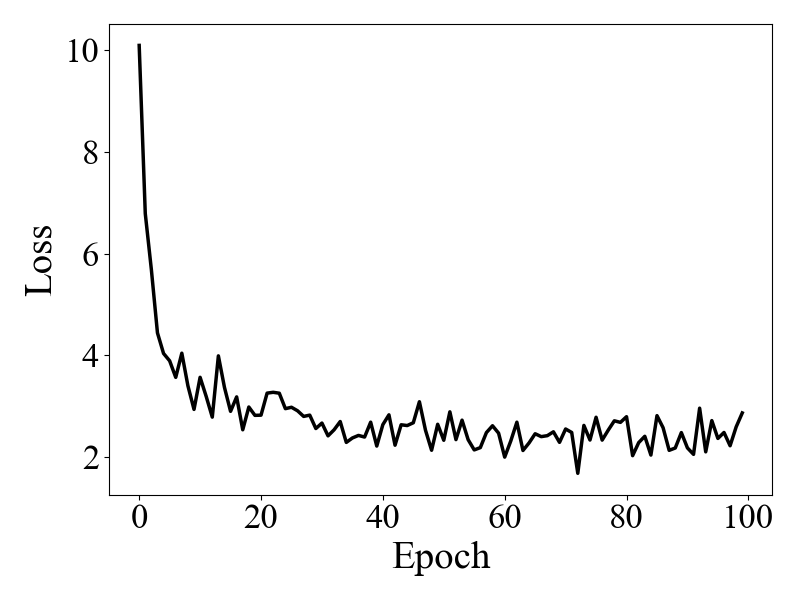}  % 设置子图宽度为33%
        % \label{subfig:loss_a}
    }
    % \hspace{0.01\linewidth}  % 控制子图之间的间距
    \subfigure[Joint Optimization Loss of HouseTwenty]{
        \centering
        \includegraphics[width=0.46\linewidth]{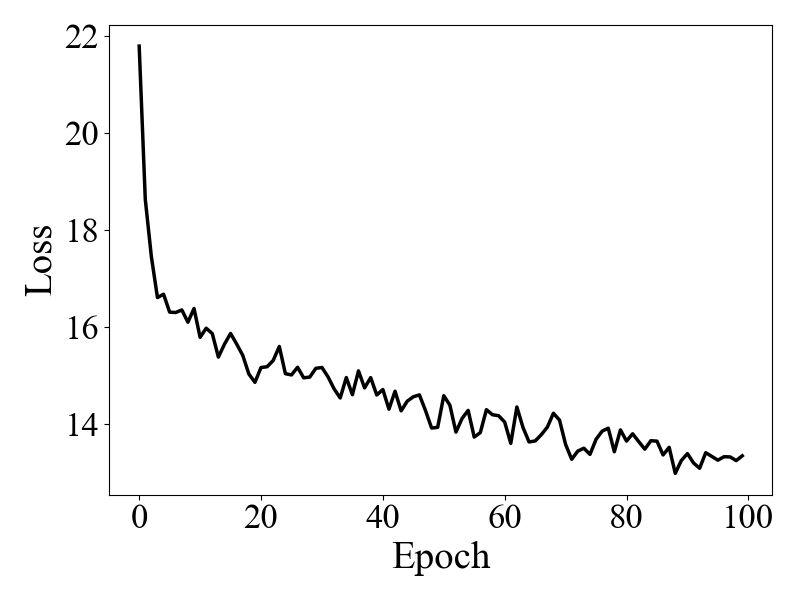}  % 设置子图宽度为33%
        % \label{subfig:loss_b}
    }
    % \hspace{0.01\linewidth}  % 控制子图之间的间距
    \caption{
        Convergence of Loss Functions During Training Process.
    }
    % \label{fig:example} %\ref{subfig:parameter_c}
\end{figure}
Figure 6 shows the convergence of the loss functions during the training process of FCACC on the FacesUCR and HouseTwenty datasets. It can be observed that the losses in both the pre-training phase and the joint optimization phase decrease rapidly and eventually stabilize. This indicates that the model effectively reduced the loss in both stages of training, achieving good convergence.
% \begin{figure}[ht]
%     \centering
%     \includegraphics[width=1\linewidth]{figures/loss.png}
%     \caption{Convergence of the loss functions during the training process.}
%     \label{fig:enter-label}
% \end{figure}

\subsubsection{Change in Clustering Performance During Training Process}
Figure 7 shows the trend of NMI and RI metrics during the training process of FCACC on the FacesUCR and HouseTwenty datasets, with the first 100 epochs representing the pre-training phase and the subsequent 100 epochs representing the joint optimization phase. 

\begin{figure}[!ht]
    \centering
    \subfigure[NMI of FacesUCR]{
        \centering
        \includegraphics[width=0.46\linewidth]{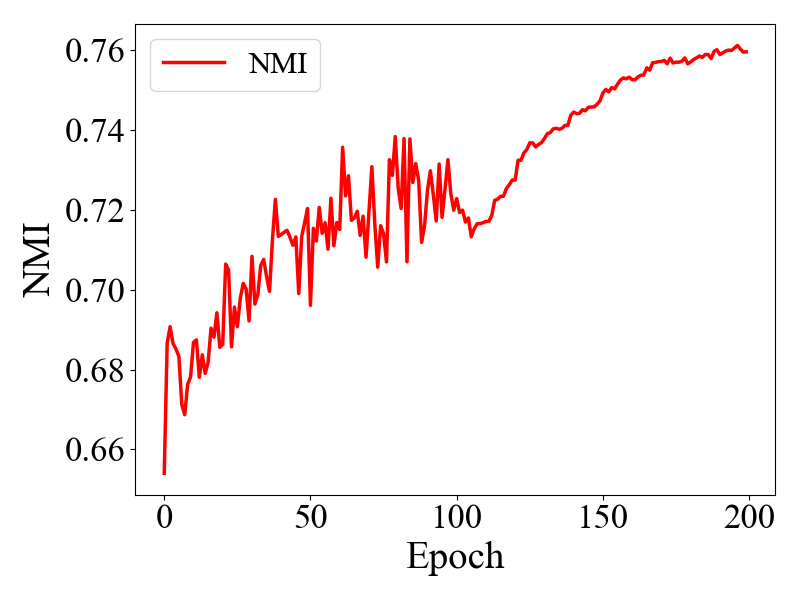}  % 设置子图宽度为33%
        % \label{subfig:loss_c}
    }
    \subfigure[RI of FacesUCR]{
        \centering
        \includegraphics[width=0.46\linewidth]{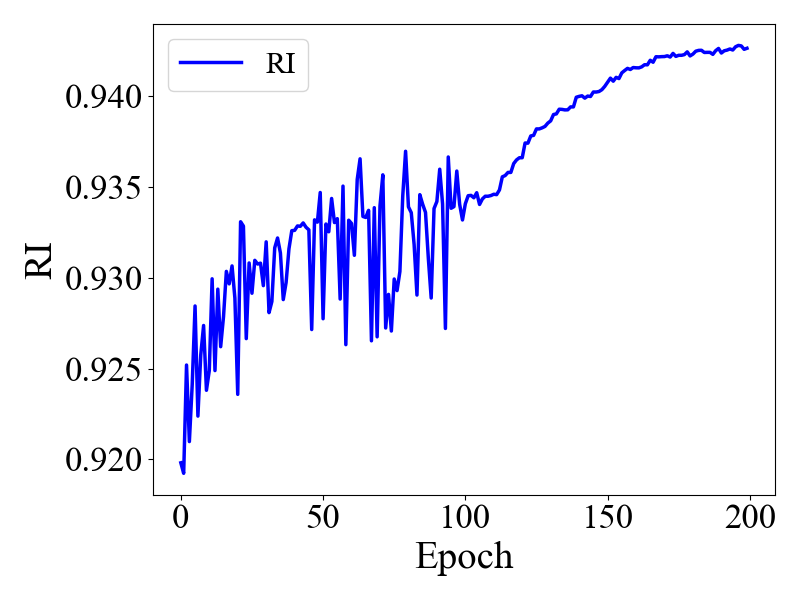}  % 设置子图宽度为33%
        % \label{subfig:loss_d}
    }
    \subfigure[NMI of HouseTwenty]{
        \centering
        \includegraphics[width=0.46\linewidth]{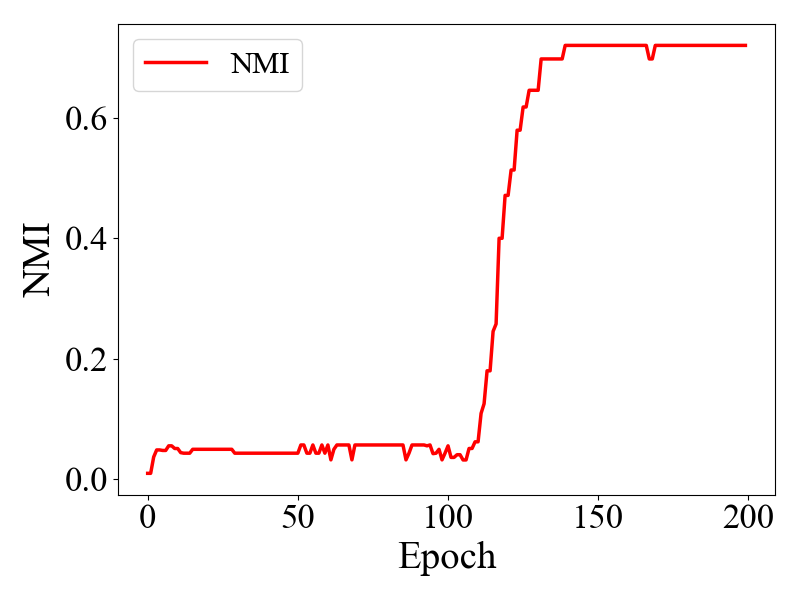}  % 设置子图宽度为33%
        % \label{subfig:loss_a}
    }
    % \hspace{0.01\linewidth}  % 控制子图之间的间距
    \subfigure[RI of HouseTwenty]{
        \centering
        \includegraphics[width=0.46\linewidth]{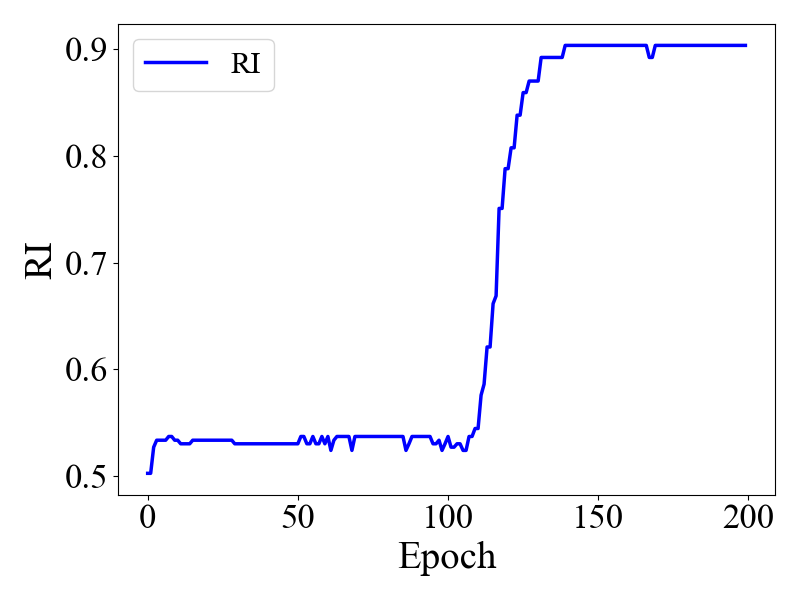}  % 设置子图宽度为33%
        % \label{subfig:loss_b}
    }
    \caption{Variation in Clustering Performance During Training Process.
    }
    % \label{fig:example} %\ref{subfig:parameter_c}
\end{figure}

On the FacesUCR dataset, the NMI and RI fluctuated significantly during the pre-training phase, likely due to the conflict between contrastive learning and the clustering objective. During the joint optimization phase, the introduction of the cluster-awareness module effectively alleviated these fluctuations and led to a steady increase in both NMI and RI. On the HouseTwenty dataset, there was little improvement in NMI and RI during the pre-training phase, indicating that the pre-training phase struggled to capture cluster structure features. In the joint optimization phase, NMI and RI quickly rose and stabilized, validating the effectiveness of the joint optimization.

Overall, Figure 7 not only demonstrates the steady improvement in performance during the training process but also clearly illustrates the different contributions of the pre-training and joint optimization phases to model performance. In the pre-training phase, the model primarily learns general feature representations through contrastive learning, while in the joint optimization phase, the model further enhances clustering performance by introducing cluster awareness and dynamic optimization mechanisms. This two-phase design allows the model to steadily learn features in the initial stage and strengthen the clustering objective in the later stage, adapting to the complexity of different datasets and fully validating the effectiveness of this framework.

\subsection{Visualization}
\begin{figure*}[!ht]
    \centering
    \subfigure[t-SNE Visualization of ShapeletSim Before Training]{
        \centering
        \includegraphics[width=0.46\linewidth]{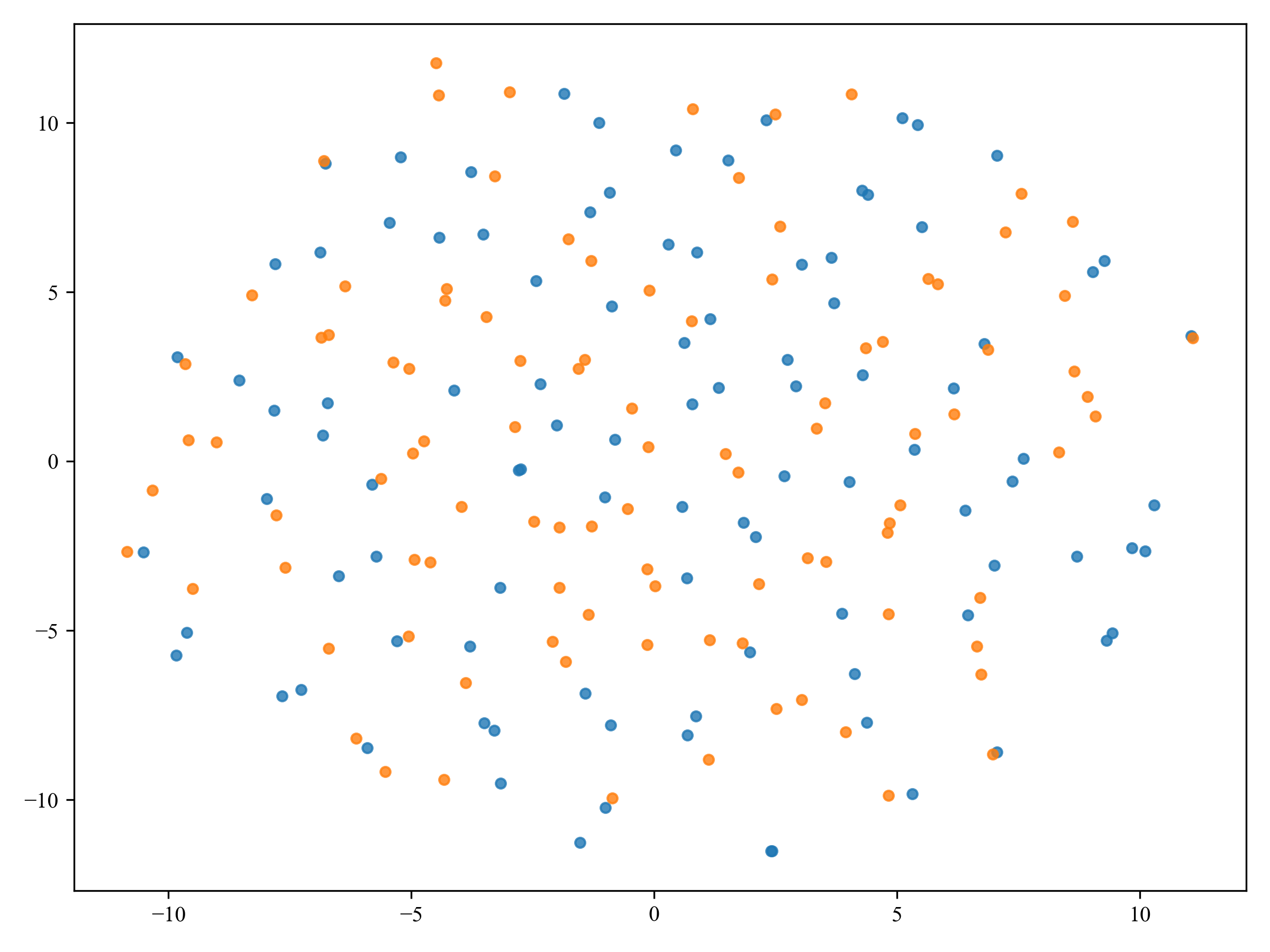}  % 设置子图宽度为33%
        \label{subfig:tsne_a}
    }
    \subfigure[ShapeletSim t-SNE Visualization After Training]{
        \centering
        \includegraphics[width=0.46\linewidth]{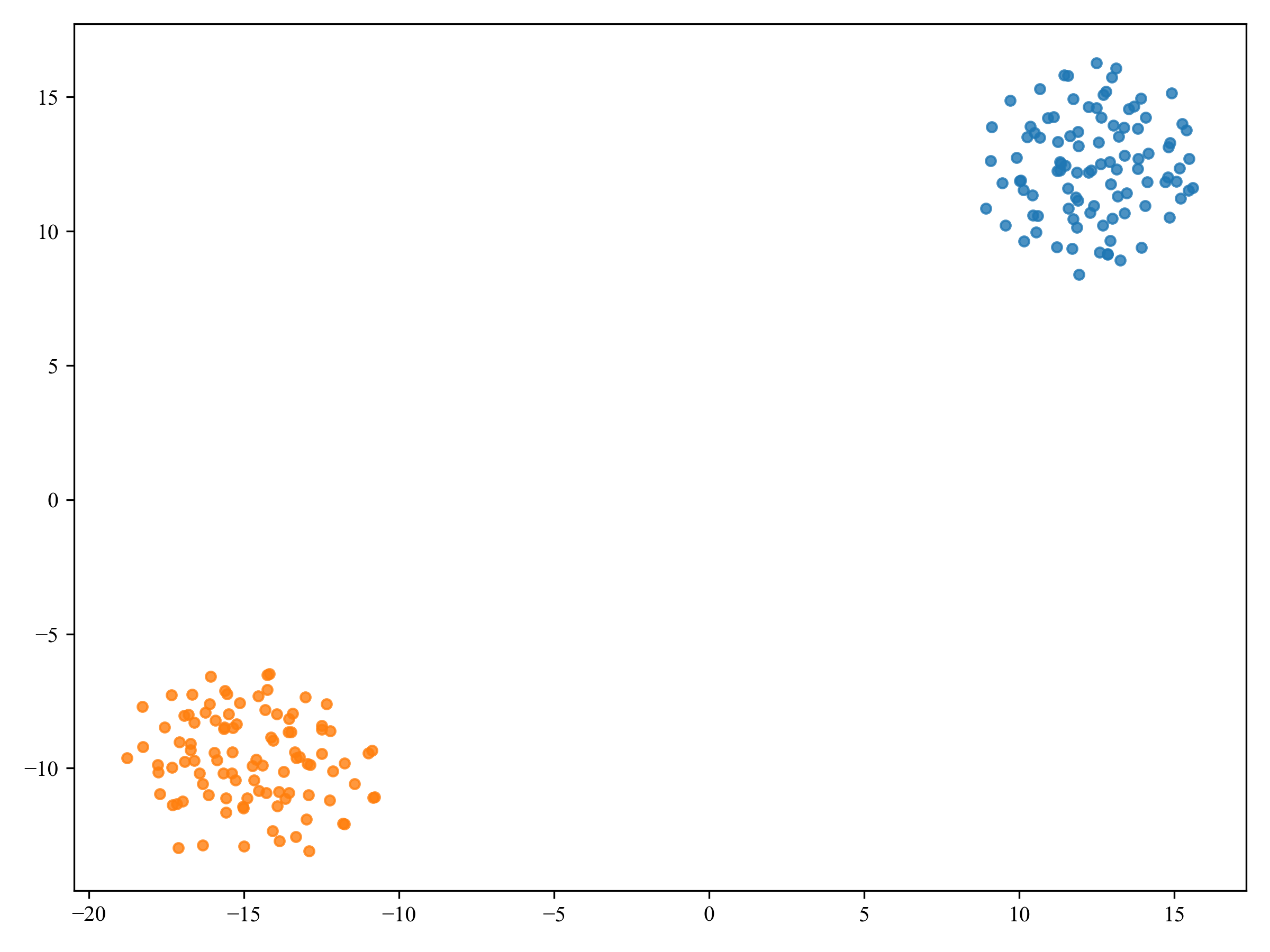}  % 设置子图宽度为33%
        \label{subfig:tsne_b}
    }
    \subfigure[t-SNE Visualization of SyntheticControl Before Training]{
        \centering
        \includegraphics[width=0.46\linewidth]{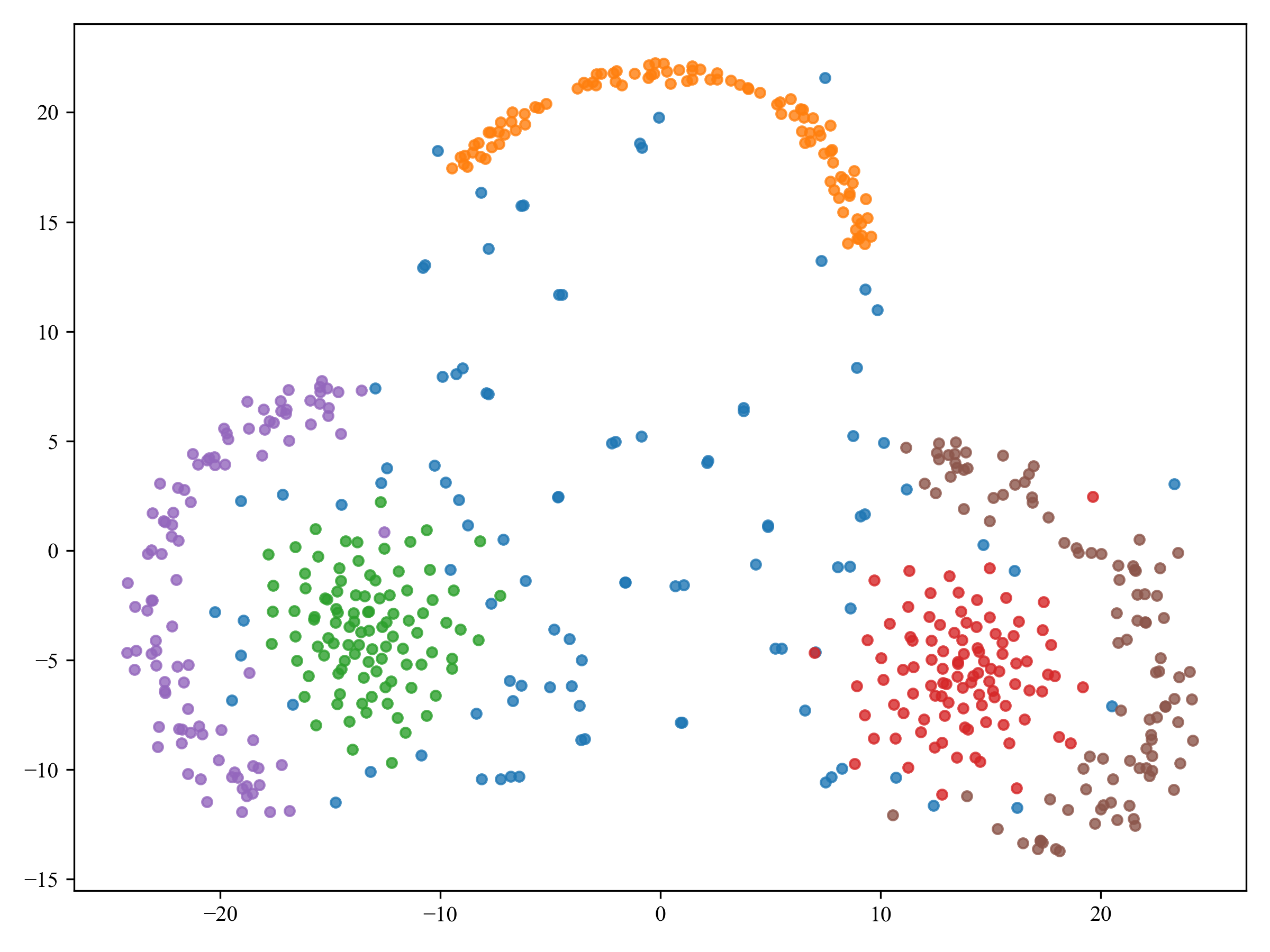}  % 设置子图宽度为33%
        \label{subfig:tsne_c}
    }
    \subfigure[SyntheticControl t-SNE Visualization After Training]{
        \centering
        \includegraphics[width=0.46\linewidth]{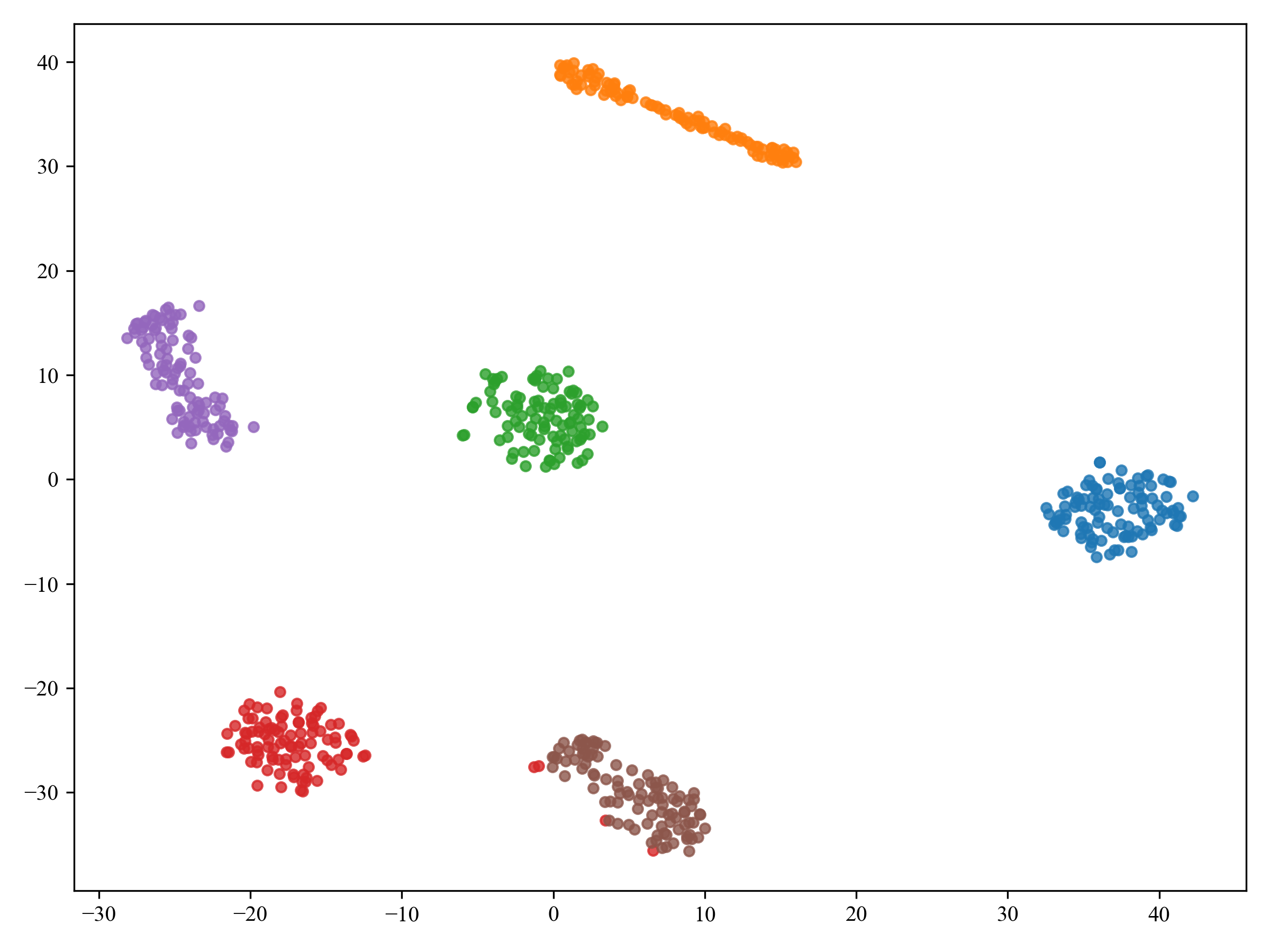}  % 设置子图宽度为33%
        \label{subfig:tsne_d}
    }
    \caption{Comparison of Sample Representation Distributions Before and After the Training.
    }
    % \label{fig:example} %\ref{subfig:parameter_c}
\end{figure*}
To further validate the effectiveness of the model, we perform t-SNE visualization analysis on the sample representation distribution.
Figure 8 illustrates the distribution change of sample representations in the embedding space before and after model training, using t-SNE visualization on the ShapeletSim and SyntheticControl datasets. Before training, the sample distribution is more dispersed and chaotic, with blurry boundaries between different clusters, making it difficult to form distinct clustering structures. After training, the samples gradually cluster together to form clear cluster structures, with increased intra-cluster tightness and significantly enhanced inter-cluster boundaries. This result indicates that the model effectively improves the discriminability of feature representations and clustering performance through joint optimization, aligning the sample distribution more closely with the clustering objective.
% \begin{figure}[ht]
%     \centering
%     \includegraphics[width=1\linewidth]{figures/tSNE.png}
%     \caption{Comparison of sample representation distributions before and after training.}
%     \label{fig:enter-label}
% \end{figure}

\section{Conclusion}
\label{sec:conclusion}
This paper proposes a fuzzy cluster-aware contrastive clustering method (FCACC), which aims to jointly optimize representation learning and clustering tasks for time series data. FCACC introduces a three-view data augmentation strategy to capture diverse temporal characteristics, a cluster-aware hard negative sample generation strategy to enhance discriminative power, and a cluster-awareness generation module to guide the joint optimization of representation learning and clustering objectives. Experiments on 40 benchmark datasets show that FCACC outperforms existing methods in overall performance, while exhibiting strong robustness to complex data patterns. Future work could explore extending FCACC to other domains, such as multi-modal data or streaming time series.

\ifCLASSOPTIONcaptionsoff
    \newpage
\fi

% balance the columns on the last page
% \IEEEtriggeratref{8}
% \IEEEtriggercmd{\enlargethispage{-5in}}

% references section
\bibliographystyle{./bibliography/IEEEtran}
%\bibliography{./bibliography/IEEEabrv,./bibliography/mybibs}
% Generated by IEEEtran.bst, version: 1.14 (2015/08/26)

% biography section
\vspace{-1 mm}
\begin{IEEEbiography}[{\includegraphics[width=1in,height=1.25in,clip,keepaspectratio]{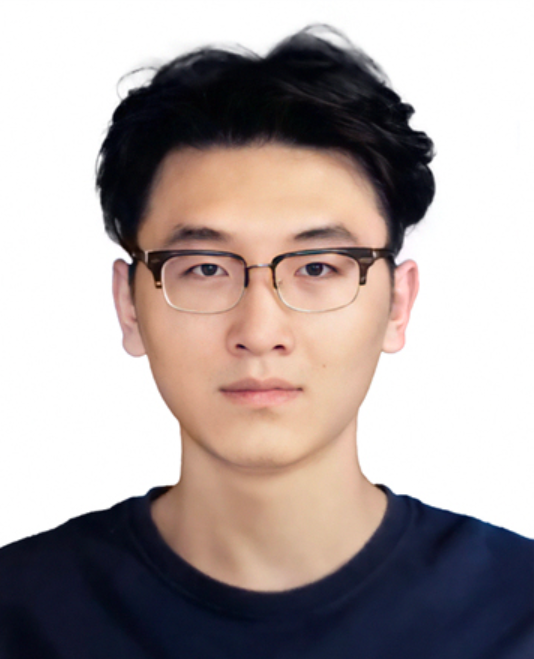}}]
    {Congyu Wang} received the bachelor's degree from Jinling Institute of Technology, Nanjing, China, in 2022. He is currently pursuing the master's degree with the School of Computer Science and Technology, Jiangsu Normal University, Xuzhou, China. 
    
    His research interests include deep learning, clustering, and time-series analysis.
\end{IEEEbiography}

\vspace{-1 mm}
\begin{IEEEbiography}[{\includegraphics[width=1in,height=1.25in,clip,keepaspectratio]{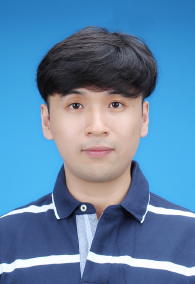}}]
    {Mingjing Du} (Member, IEEE) received the Ph.D. degree in computer science from the China University of Mining and Technology, Xuzhou, China, in 2018.

    He is currently an Associate Professor with the School of Computer Science and Technology, Jiangsu Normal University, Xuzhou, China. His current research interests include cluster analysis and three-way decisions.
    For more information, see https://dumingjing.github.io/
\end{IEEEbiography}

\vspace{-1 mm}
\begin{IEEEbiography}[{\includegraphics[width=1in,height=1.25in,clip,keepaspectratio]{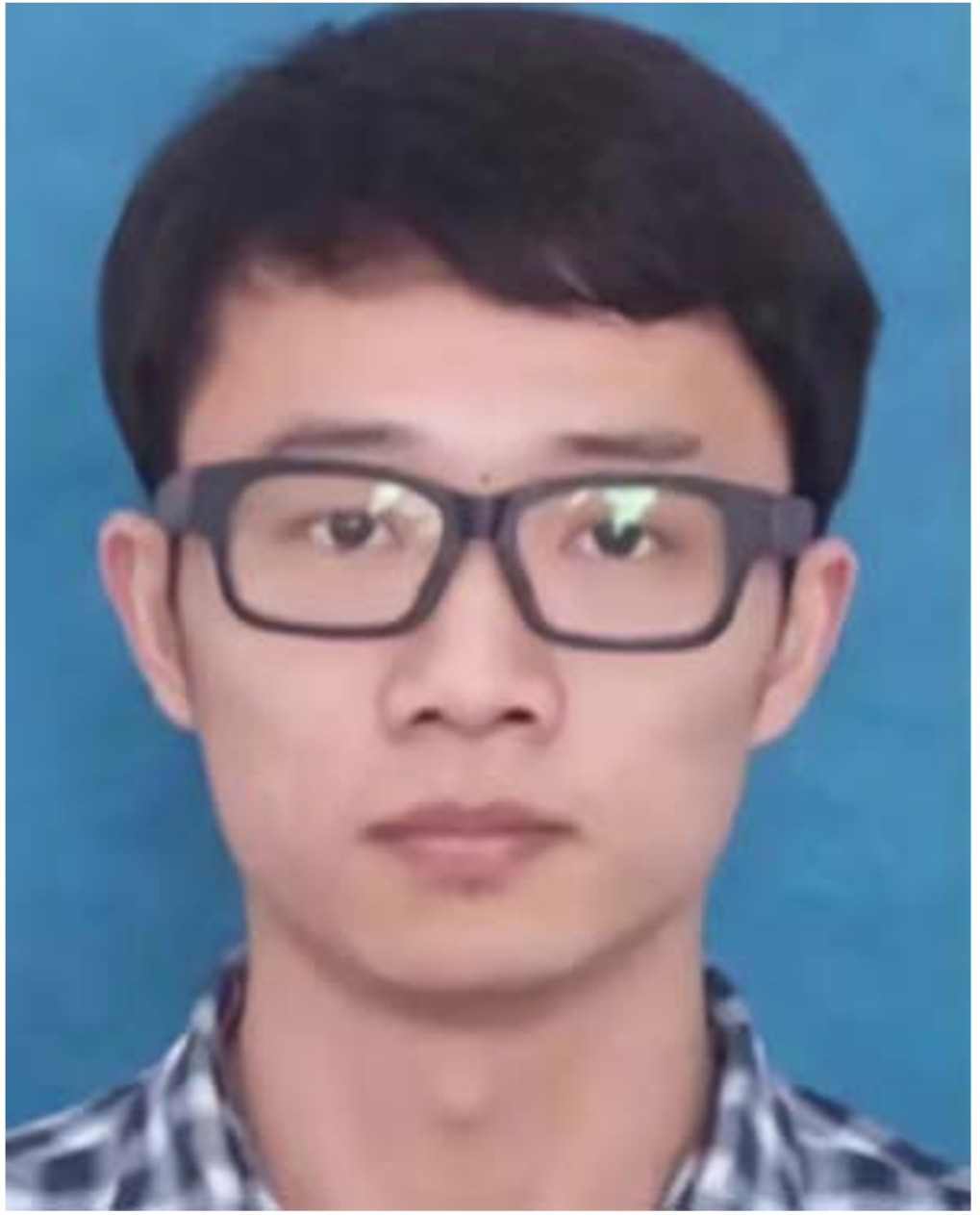}}]
    {Xiang Jiang} received the Ph.D. degree in signal and information processing from Beijing JiaoTong University, Beijing, China, in 2022. He is currently an Assistant Professor with the SchoolofComputerScienceandTechnology,Jiangsu Normal University, Xuzhou, China. 
    
    His research interests include image forensic and computer vision.
\end{IEEEbiography}

\vspace{-1 mm}
\begin{IEEEbiography}[{\includegraphics[width=1in,height=1.25in,clip,keepaspectratio]{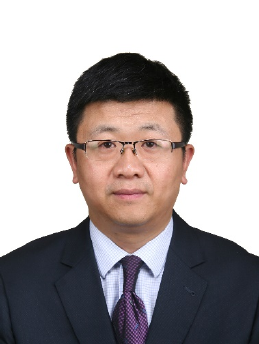}}]
    {Yongquan Dong} received the Ph.D. degree in computer science from Shandong University, Jinan, China, in 2010.

    He is currently a Professor with the School of Computer Science and Technology, Jiangsu Normal University, Xuzhou, China. His research interests include web information integration and web data management.
\end{IEEEbiography}

% \vfill
% \enlargethispage{-5in}

\end{document}